%% file: arxiv.tex
\documentclass{article}

\usepackage[utf8]{inputenc} % Allow UTF-8 input
\usepackage[T1]{fontenc}    % Use 8-bit T1 fonts

\usepackage{graphicx}       % For including graphics
\usepackage{subcaption}     % For subfigures

\usepackage{booktabs}       % Professional-quality tables

\usepackage{amsmath, amsfonts, amssymb, amsthm, mathtools}

\usepackage[numbers]{natbib} % or [authoryear] if you prefer author-year citations
\usepackage{algorithm}
\usepackage{algorithmic}

\usepackage[colorlinks=true, linkcolor=blue, citecolor=blue, urlcolor=blue]{hyperref}

\usepackage{multicol}

\usepackage{nicefrac}       % Compact symbols for 1/2, etc.
\usepackage{microtype}      % Microtypography
\usepackage{xcolor}         % Colors
\usepackage{cleveref}       % Intelligent cross-referencing

\usepackage[preprint]{arxiv}

\input{math_commands.tex}

\title{Just How Flexible are Neural Networks in Practice?}

\author{%
  Ravid Shwartz-Ziv\thanks{Authors contributed equally.} \\
  New York University \\
  $\texttt{ravid.shwartz.ziv@nyu.edu}$ 
  \And
  Micah Goldblum$^{*}$ \\
  New York University \\
  \And
  Arpit Bansal\\
  University of Maryland \\
  \And
  C. Bayan Bruss\\
  Capital One\\
  \And
  Yann LeCun\\
  New York University\\
  Meta AI, FAIR
  \And
  Andrew Gordon Wilson\\
  New York University \\
}

\begin{document}

\maketitle

\begin{abstract}

It is widely believed that a neural network can fit a training set containing at least as many samples as it has parameters, underpinning notions of \emph{overparameterized} and \emph{underparameterized} models.  In practice, however, we only find solutions accessible via our training procedure, including the optimizer and regularizers, limiting flexibility.  Moreover, the exact parameterization of the function class, built into an architecture, shapes its loss surface and impacts the minima we find. In this work, we examine the ability of neural networks to fit data in practice.  Our findings indicate that: (1) standard optimizers find minima where the model can only fit training sets with significantly fewer samples than it has parameters; (2) convolutional networks are more parameter-efficient than MLPs and ViTs, even on randomly labeled data; (3) while stochastic training is thought to have a regularizing effect, SGD actually finds minima that fit more training data than full-batch gradient descent; (4) the difference in capacity to fit correctly labeled and incorrectly labeled samples can be predictive of generalization; (5) ReLU activation functions result in finding minima that fit more data despite being designed to avoid vanishing and exploding gradients in deep architectures.
\end{abstract}

\section{Introduction}

Neural networks are often assumed to be capable of fitting about as many samples as they have parameters \citep{zhang2016understanding, arpit2017closer, dziugaite2017computing}.  This intuition can be most easily understood through linear regression, where a regressor with more coefficients than training samples forms an underdetermined linear system of equations and can therefore precisely fit any function of the training points.  For example, consider that for any training set $\{(x_i,y_i)\}_{i=0}^{n}$ with $n\leq d$, there exist parameters $\{a_i\}$ such that $f(x)=\sum_{j=0}^{d} a_jx^j$ has $f(x_i)=y_i$ for all $i$ as long as no two training points contain the same input but assigned different labels.

The theory underlying neural networks is significantly more complicated.  A variety of approximation theories bound the number of parameters or hidden units required by a neural network architecture to approximate a certain function class on its domain, which is typically infinite \citep{hornik1989multilayer, barron1992neural, mhaskar2016deep}.  On finite domains, namely a training set, \emph{overparameterized} neural networks with many more parameters than training samples can easily fit randomly labeled data, raising questions regarding how such flexible models can still generalize to new unseen test data \citep{zhang2016understanding}.

In this work, we step back and ask just how flexible neural networks really are in practice.  Although neural networks are theoretically capable of universal function approximation \citep{hornik1989multilayer}, in practice we train models with limited capacity and only find optima during training that are accessible via our training procedure, often leading to significantly reduced flexibility as suboptimal local minima exist \citep{goldblum2020truth}.  How much data we can fit depends on factors like the nature of the data itself, model architecture, size, optimizer, and regularizers. 
In this work, we measure the capacity of models to fit data under realistic training loops, and we examine the effects of various features like architectures and optimizers on the number of training samples a model can fit in practice.  Our findings are summarized as follows:

\begin{itemize}
\item The optimizers typically used for training neural networks often find minima where the model can only perfectly fit training sets with far fewer samples than model parameters.  This observation calls into question whether we actually find overfitting local minima in practice, contrary to conventional wisdom.
\item Convolutional architectures (CNNs) are known to generalize better than multi-layer perceptrons on computer vision problems due to their strong inductive bias for spatial relationships and locality. However, we find that CNNs are actually more parameter efficient on randomly labeled data as well, indicating that their superior capacity to fit data does not result from superior generalization alone.
\item The ability of a neural network to fit many more correctly labeled samples than incorrectly labeled samples is predictive of generalization.
\item ReLU activation functions enable fitting more training samples than sigmoidal activations after successfully finding minima using models with each activation function, even though ReLU nonlinearities were introduced to neural networks to prevent vanishing and exploding gradients in deep neural networks with many layers.
\item SGD is thought to have a regularizing effect that improves generalization, yet we find that SGD actually enables fitting more training samples than full-batch gradient descent.
\end{itemize}

\section{Related Work}

\textbf{Approximation theory.} A primary area of early deep learning theory focused on upper bounding the number of parameters or neurons required to well-approximate functions in a particular class, for example uniform approximation of continuous functions on a compact domain \citep{hornik1989multilayer}.  Such approximation theories typically focus on arbitrary compact sets or data on a well-behaved manifold \citep{hornik1989multilayer, shaham2018provable}.  The resulting upper bounds are often proved constructively, and the constructions may be specific to a particular neural network architecture, often very shallow networks with only a few layers, limiting their generality.  We focus on neural network flexibility on the training set, empirically measuring the parameters needed to fit real data in practice, rather than theoretical bounds.  This methodology allows us to try any architecture or to inspect the influence of optimizers, and it measures quantities that actually impact neural networks.

\textbf{Overparameterized neural networks and generalization.}  Early generalization theories predicted that highly constrained models which fit their training data yet fail to fit randomly labeled data (\emph{i.e.} have low Rademacher complexity or VC-dimension) can generalize to new unseen test data \citep{vapnik1991principles, bartlett2002rademacher}.  However, these theories fail to account for the exceptional generalization behavior of neural networks since they are often highly flexible and overparameterized, leading to vacuous error bounds \citep{zhang2016understanding}.  Recent work on PAC-Bayes generalization theory explains that highly flexible and overparameterized models can generalize well as long as they assign disproportionate prior mass to parameter vectors which fit the training data \citep{dziugaite2017computing, lotfi2022pac, lotfi2023non}.  Related empirical works explain why neural network inductive bias and consequently generalization can actually benefit from overparametrization \citep{huang2019understanding, chiang2022loss, maddox2020rethinking}.  We will see in our own experiments that whereas the Rademacher complexity of neural networks is extremely high, they can fit many more correctly labeled samples than randomly labeled ones in practice, and this gap predicts generalization. Nakkiran et. al \citep{nakkiran2021dee1} use the data-fitting capacity of neural networks to understand the double-descent phenomenon. They train networks with many fewer or many more parameters than the number of samples they can fit, and they study the impact of such over- and underparameterization on generalization.  In contrast, we are interested in what influences that capacity to fit data itself.

\section{Preliminaries}

\textbf{Quantifying capacity.} While it is straightforward to determine the number of samples a linear regression model can fit by counting its parameters, neural networks present a more complicated story. Our goal is to measure the neural network's capacity to fit real data using realistic training routines. This metric should satisfy three essential criteria: (1) it must measure the real-world capacity to fit data, enabling us to evaluate the effects of optimizers and regularizers; (2) it should be sensitive to the training dataset, meaning it should reflect the capacity to fit different types of data or data with specific labeling characteristics; and (3) it must be feasible to compute.

To that end, we adopt the \emph{Effective Model Complexity} (EMC) metric \citep{nakkiran2021dee1}, which estimates the largest sample size that a model can perfectly fit.  We apply this metric across various data types, including those with random or semantic labels or even random inputs.

Calculating EMC involves an iterative approach for each network size. Initially, we train the model on a small number of samples. If it achieves 100\% training accuracy after training, we re-initialize and train on a larger set of randomly chosen samples. We iteratively perform this process, incrementally increasing the sample size each time until the model no longer fits all training samples perfectly. The largest sample size where the model still achieves perfect fitting is taken as the network's EMC. It is important to note that the initialization and data subsets we sample on each iteration are independent of those from previous iterations, ensuring that our capacity evaluation remains unbiased. Where the network did not successfully reach 100\% training accuracy, we re-run training three more times with different random seeds to ensure that inability to fit all samples was not a fluke. Furthermore, we also tried performing all analysis instead with a relaxed requirement that the network fit $98\%$ of its training data, which did not significantly affect results.

While it is possible to artificially prevent models from fitting their training set by under-training, confounding any study of capacity to fit data, we ensure that all training runs reach a minimum of the loss function by imposing three conditions: first, the norm of the gradients across all samples must fall below a pre-defined threshold; second, the training loss should stabilize; third, we check for the absence of negative eigenvalues in the loss Hessian to confirm that the model has indeed reached a minimum rather than a saddle point.  In \Cref{ref:edc}, we detail our method for computing the EMC as well as how we enforce the above three conditions.

In contrast to Nakkiran et al. \citep{nakkiran2021dee1}, we validate that each model reaches a minimum during optimization by ensuring the absence of negative singular values of the loss Hessian.  This step is important given that we train models of various architectures and sizes, so we want to prevent under-training from being a confounding variable.

\textbf{Underparameterization and overparameterization.} Linear models are described as \emph{underparameterized} when they have fewer parameters than training samples and \emph{overparameterized} when they have more parameters than training samples.  This threshold determines when a linear regression model can fit any labeling of its data, and it often coincides with the transition to strict convexity when a linear model has a unique optimal parameter vector.  Neural networks behave differently than linear regression models; their loss function is non-convex and can have multiple minima even when training sets are large.  Moreover, it is unclear exactly how many parameters a neural network needs to fit its training set in practice.  We will use EMC to investigate the latter quantity.

\textbf{The differences between capacity, flexibility, expressiveness, and complexity.}
These terms are used in numerous ways, sometimes interchangeably and sometimes distinctly.  For example, Rademacher complexity and VC-dimension are notions of complexity typically associated with flexibility, whereas the PAC-Bayes notion of complexity is information-theoretic and instead measures compression.  Expressiveness can be used to described the breadth of an entire hypothesis class, that is all the functions that a model can express across all possible parameter settings.  Approximation theories measure the expressiveness of a hypothesis class by the existence of elements of this class which well-approximate functions of a specified type. We will abstain from using the terms ``expressiveness'' and ``complexity'' when describing EMC to avoid confusion, and we will use ``capacity'' and ``flexibility'' when referring to a model's ability to fit data in practice.

\textbf{Factors influencing the EMC.}
Unlike VC-dimension or expressiveness concepts in approximation theories, EMC depends not only on the hypothesis class but on every aspect of neural network training, from optimizers and regularizers to the specific parameterization induced by the model's architecture. Choices in architectural design and training algorithms influence the loss of surface geometry, thereby affecting the accessibility of certain solutions.

\section{Experimental Setup}

We conduct a comprehensive dissection of the factors influencing neural network flexibility. To this end, we consider a variety of datasets, architectures, and optimizers.

\subsection{Datasets}

We conduct experiments on a variety of datasets, including vision datasets like MNIST \citep{deng2012mnist}, CIFAR-10, CIFAR-100 \citep{krizhevsky2009learning}, and ImageNet \citep{deng2009imagenet}, as well as tabular datasets like Forest Cover Type \citep{BLACKARD1999131}, Adult Income \citep{misc_adult_2}, and the Credit dataset \citep{creditcard_dataset}. Due to the small size of these datasets, we also use larger synthetic datasets. These are generated using the Efficient Diffusion Training via Min-SNR Weighting Strategy \citep{hang2023efficient}, yielding diverse ImageNet-quality samples at a resolution of $128 \times 128$. Specifically, we create ImageNet-20MS, containing 20 million samples across ten classes. Unless otherwise specified, the main text describes results on ImageNet-20MS, while the appendix contains results on additional datasets. We omit data augmentations to avoid confounding effects.

\subsection{Models}

We evaluate the flexibility of diverse architectures, including Multi-Layer Perceptrons (MLPs), CNNs like ResNet \citep{he2016deep} and EfficientNet \citep{tan2019efficientnet}, and Vision Transformers (ViTs) \citep{dosovitskiy2020image}. We systematically adjust the width and depth of these architectures. For MLPs, we either increase the width by adding neurons per layer while keeping the number of layers constant or increase the depth by adding more layers while keeping the number of neurons per layer constant. For naive CNNs, we employ multiple convolutional layers followed by a constant-sized fully connected layer, varying either the number of filters per layer or the total number of layers. For ResNets, we scale either the number of filters or the number of blocks (depth). In ViTs, we scale the number of encoder blocks (depth), the dimensionality of patch embeddings, and self-attention (width). By default, we scale the width to control the parameter count unless stated otherwise.

\subsection{Optimizers}

We employ several optimizers, including Stochastic Gradient Descent (SGD), Adam \citep{kingma2014adam}, AdamW \citep{loshchilov2018decoupled}, full-batch Gradient Descent (GD), and the second-order Shampoo optimizer \citep{anil2021towards}. These choices let us examine how features like stochasticity and preconditioning influence the minima.  To ensure effective optimization across datasets and model sizes, we carefully tune the learning rate and batch size for each setup, omitting weight decay in all cases. Further details about our hyperparameter tuning are provided in \Cref{app:impl_details}. By default, we use SGD.

\section{The Effect of the Data on EMC}
In this section, we dissect how data properties shape neural networks flexibility and how this behavior can predict generalization.

\begin{figure}[h]
    \centering
    \begin{subfigure}[t!]{0.52\textwidth}
        \centering
        \includegraphics[width=1\linewidth]{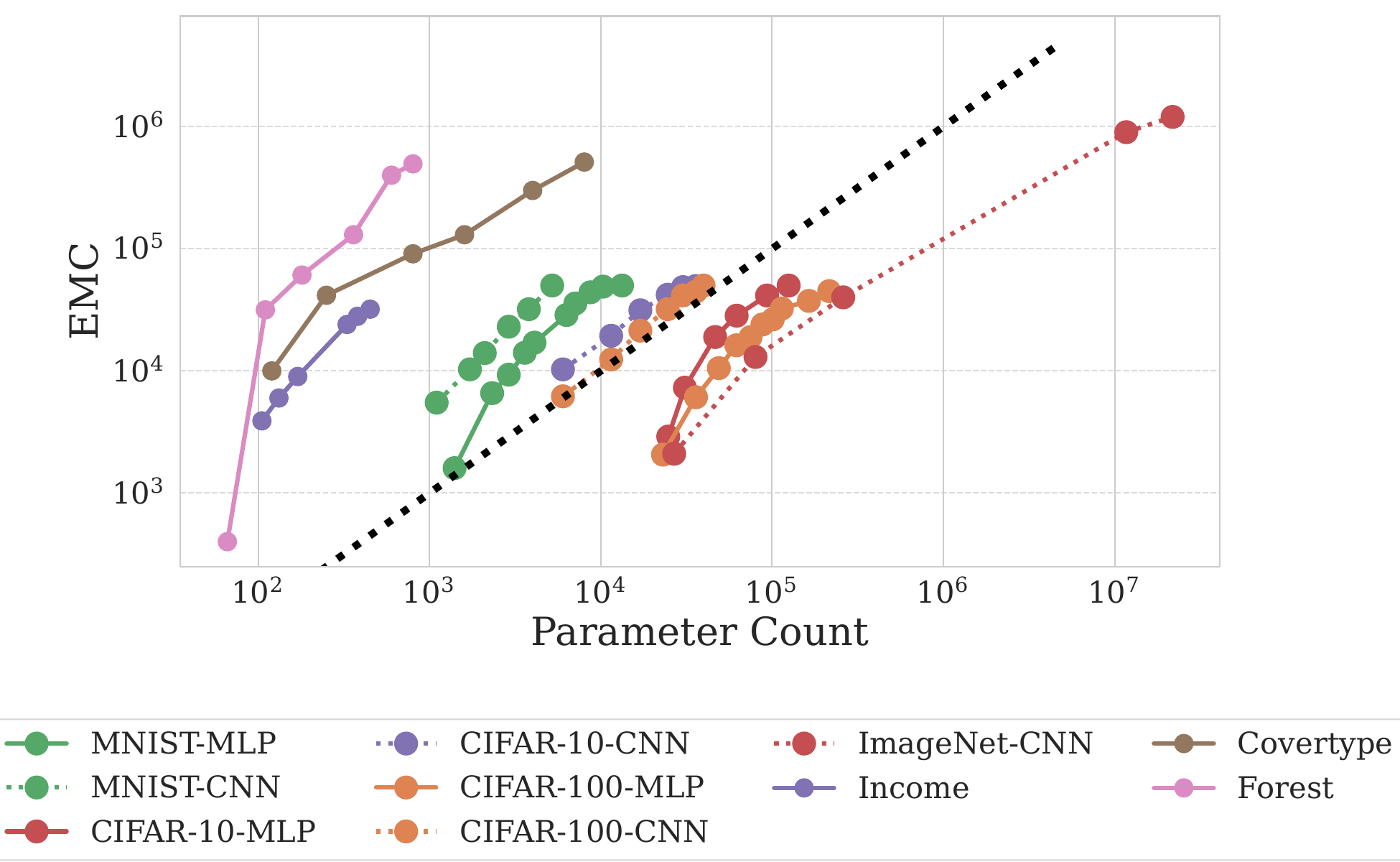}
        \label{fig:dataset_capacity}
    \end{subfigure}
    ~
    %\hfill
    \begin{subfigure}[t!]{0.46\textwidth}
        \centering
        \vspace{-6mm}
        \includegraphics[width=1\linewidth]{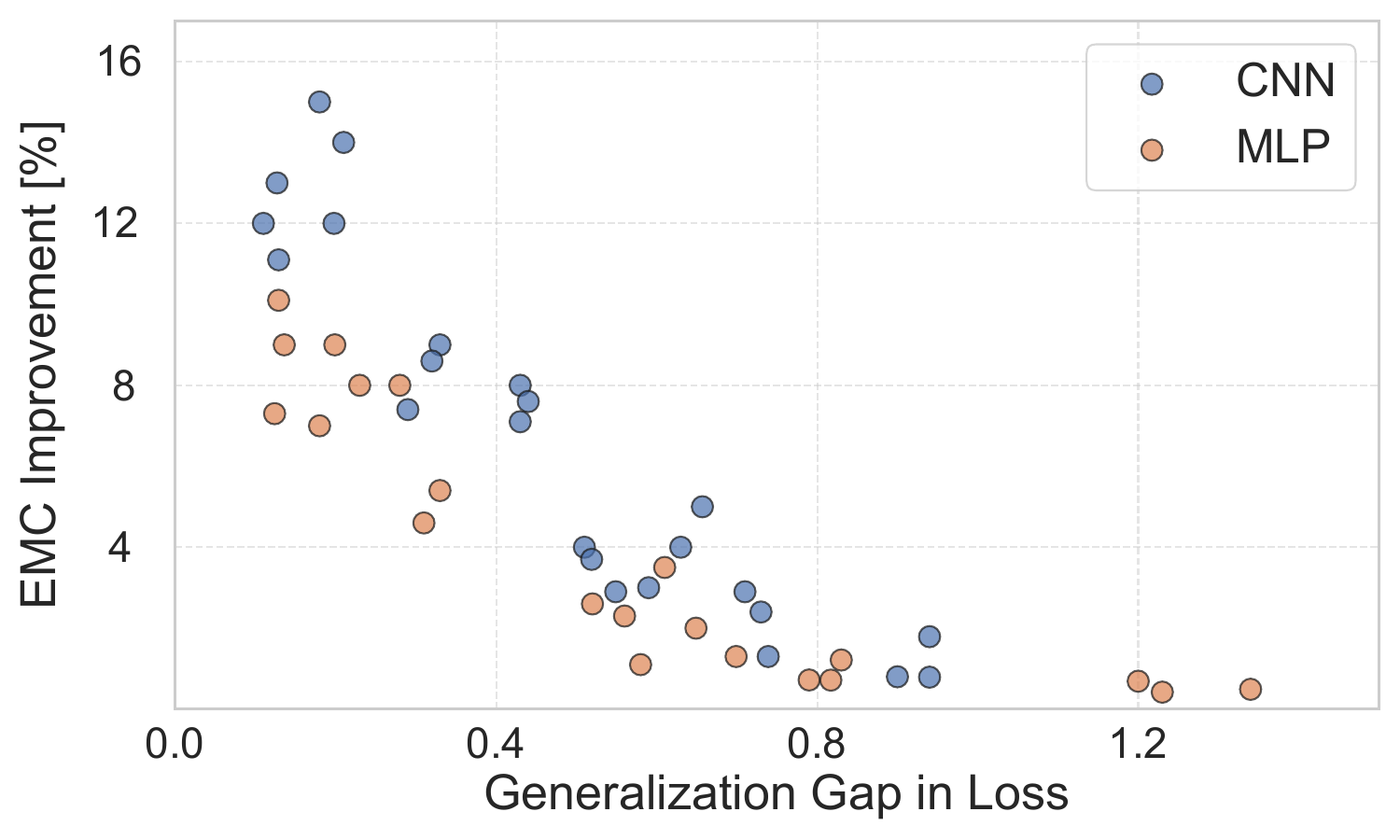}
        \label{fig:gen}
    \end{subfigure}
    \caption{ \textbf{Left: easier tasks tend to have higher EMC.} EMC across datasets and data modalities. The tabular data sets (Forest, Income, CoverType), which are easier to learn, have the highest EMC compared to vision datasets. The dashed black line is the diagonal. ImageNet is the hardest dataset to learn. \textbf{Right: the difference in EDC on the original and random labels predicts generalization.} EMC improvement as a function of the parameter count for CIFAR-100.}
    \label{fig:complete3}
\end{figure}

\textbf{Analysis of diverse datasets.}
We initiate our analysis by measuring the EMC of neural networks across various datasets and modalities. We scale a 2-layer MLP  by modifying the width of the hidden layers and a CNN by modifying the number of layers and channels, and we train models on a range of image classification (MNIST, CIFAR-10, CIFAR-100, ImageNet) and tabular (CoverType, Income, and Credit) datasets. The results reveal significant disparities in the EMC of networks trained on different data types (see \Cref{fig:complete3} (Left)). For instance, networks trained on tabular datasets exhibit higher capacity. Among image classification datasets, we observe a strong correlation between test accuracies and capacity. Notably, MNIST (where models achieve more than 99\% test accuracy) yields the highest EMC, whereas ImageNet shows the lowest, pointing to the relationship between generalization and the data-fitting capability.

Considering the variety of datasets and network architectures and the myriad differences in their EMC, the subsequent sections will explore the underlying causes of these variations. Our goal is to identify the distinct factors in the data and architectures that contribute to these observed differences in network flexibility.

\subsection{The role of inputs and labels} 
We next analyze the inductive biases of different architectures and how factors like spatial structure influence the ability of a model to fit its training data. To this end, we altering inputs and labels, measuring resulting effects. We adjust the width of MLPs and 2-layer CNNs by varying the number of neurons (MLPs) or filters (CNNs) in each layer, and we train them on ImageNet-20MS. We evaluate EMC as a function of the model's parameter count in four scenarios: semantic labels, random labels, random inputs, and inputs under a fixed random permutation. In the case of random labels, we maintain the input but sample the class labels randomly. For random inputs, we replace the original inputs with Gaussian noise, while for the permuted input, we use the same fixed permutation for all the images, breaking the spatial structure in the data.

\begin{figure*}[h]
\centering
    \includegraphics[width=1.\linewidth]{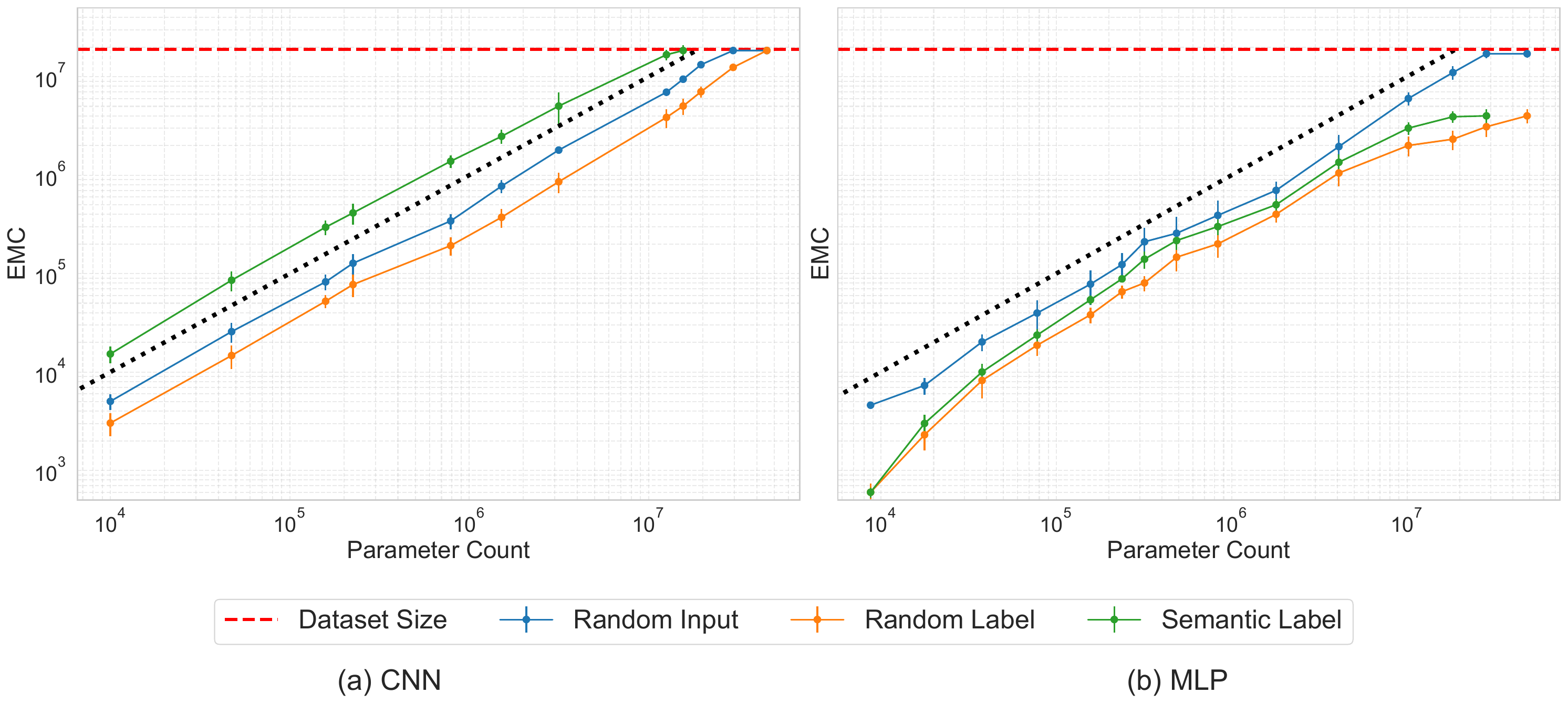}
\caption{\textbf{CNNs fit more semantically labeled samples than they have parameters due to their superior image classification inductive bias, whereas MLPs cannot.} EMC as a function of the number of parameters for semantic labels vs. random input and labels for MLPs \textbf{(a)} and CNNs \textbf{(b)}. Experiments performed on ImageNet-20MS.  Error bars represent one standard error over 5 trials.}
\label{fig:random}
%\end{figure}
\end{figure*}
\textbf{The boundary between overparameterization and underparameterization.} Linear regression models can fit at least as many samples as they have parameters, regardless of whether the labels are naturally occurring or random. The boundary between where a model has too few parameters to fit its data and where it has extra degrees of freedom is clear for linear regression. Naturally occurring labels present a more complicated scenario; for instance, if the data's labels are a linear function of the inputs, then the model can fit infinitely many samples. In \Cref{fig:random}, assigning random labels instead of real ones allows us to explore an analogous notion of the boundary between over- and under-parameterization, but in the context of neural networks.  We see here that the networks fit significantly fewer samples when assigned random labels compare to the original labels, indicating that neural networks are less parameter efficient than linear models in this setting.  Like linear models, the amount of data they can fit appears to scale linearly in their parameter count.

\textbf{The effect of high-dimensional data.} Linear models exhibit increased capacity when adding more features, primarily because their parameter count directly scales with the feature count. However, the dynamics shift when examining CNNs. In our setup, we avoid adding parameters as the data dimensionality increases by employing average pooling prior to the classification head, a standard technique for CNNs. We investigate the EMC using the ImageNet-20MS, systematically resizing input images to vary their spatial dimensions from $(16\times16)$ to $(256\times 256)$.

In contrast to linear models, we find in Appendix \Cref{fig:input_size_flexibility} that CNNs, which do not benefit from additional parameters as the input dimensionality increases, can actually fit more semantically labeled data in lower spatial dimensions.  This trend underscores a broader narrative in neural networks: CNNs, despite their intricate architectures and capacity for complex pattern recognition, tend to align better with data of lower intrinsic dimension. This observation resonates with the findings of  Pope et al. \citep{pope2020intrinsic}, who find that CNNs generally showcase enhanced generalization capabilities with data of lower intrinsic dimensionality.

\begin{figure}[t]
    \centering
   \begin{subfigure}[t!]{0.49\textwidth}
        \centering
        \vspace{-3mm}
        \includegraphics[width=1\linewidth]{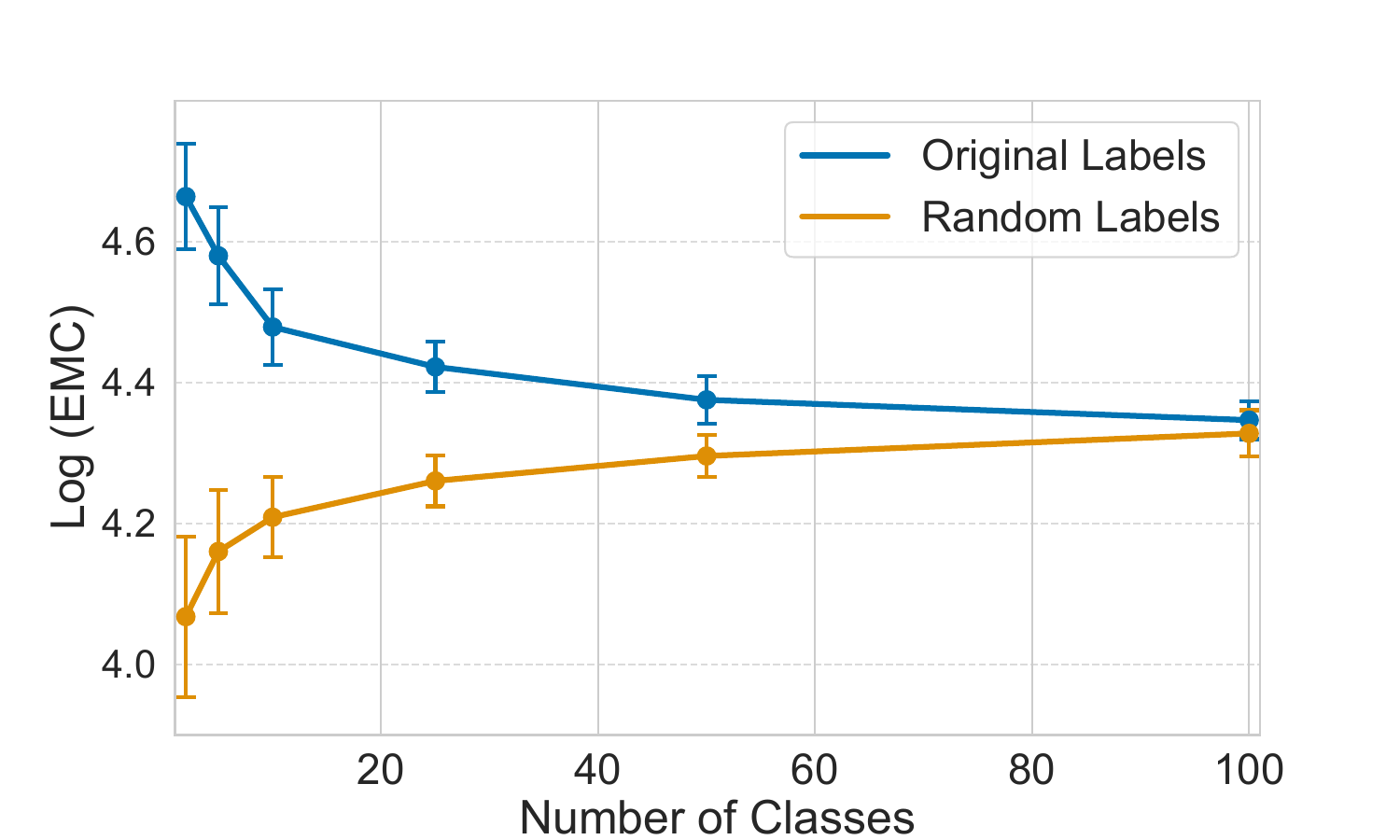}
        \vspace{-1mm}
        \caption{\textbf{More classes makes fitting data harder with semantic labels but easier with random ones.}}
        \label{fig:num_classes}
    \end{subfigure}
    \hfill
     \begin{subfigure}[]{0.49\textwidth}
        \centering
        \includegraphics[width=1\linewidth]{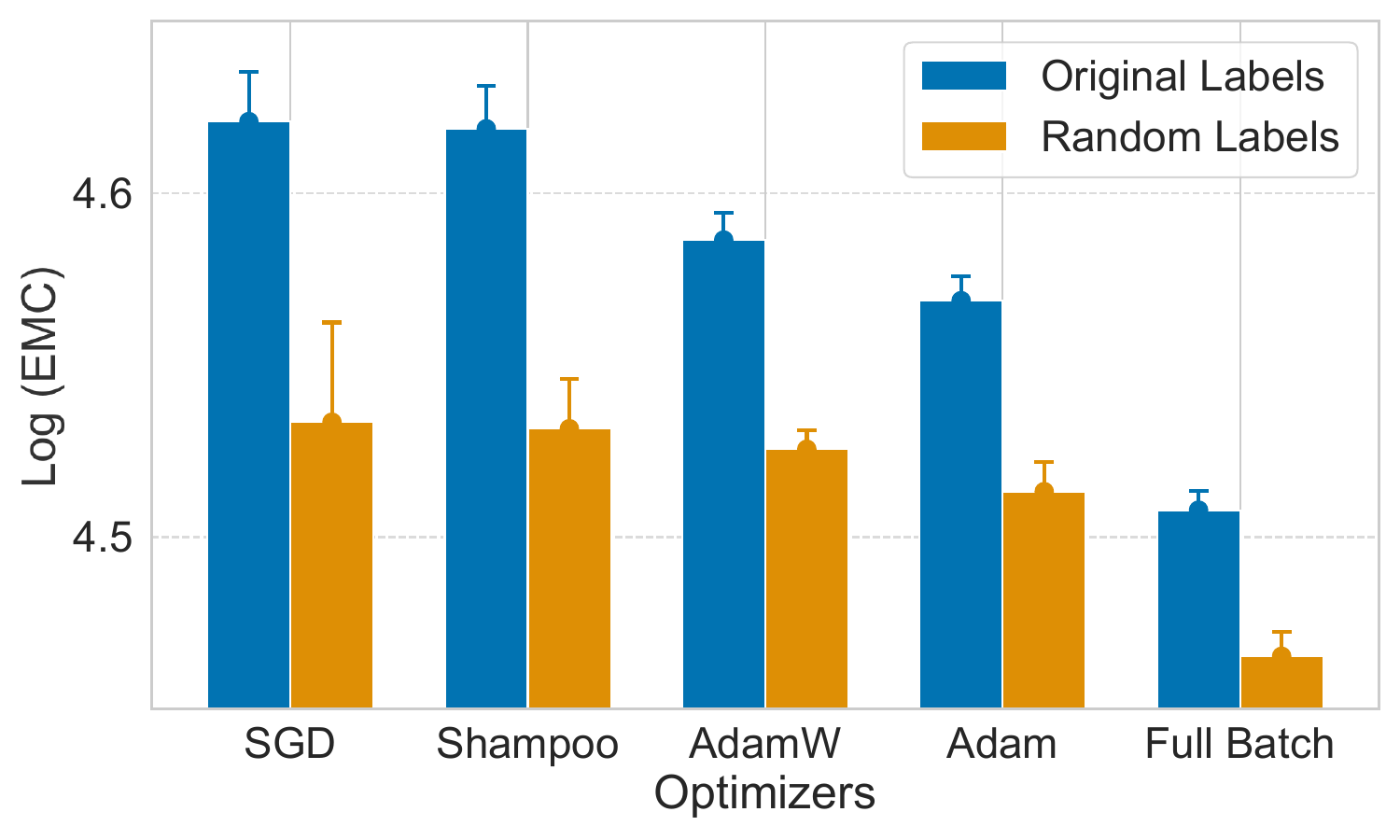}
        \caption{\textbf{SGD and Shampoo are  better for fitting with the original labels but not with random ones}}
        \label{fig:optimizers}
    \end{subfigure}
    \caption{\textbf{The effect of the number of labels and optimizers on capacity.} Average logarithm of EMC across different model sizes of CNNs on CIFAR-100 for original and random labels varying numbers of classes \textbf{(a)} and for different optimizers \textbf{(b)}. Error bars are standard error over 5 trials.}
    \label{fig:complete}
\end{figure}

\textbf{The effect of the number of classes.} In order to probe the influence of the number of classes on the EMC, we randomly merge CIFAR-100 classes to artificially decrease the number of classes while still preserving the size of the original dataset. We again consider a 2-layer CNN with various numbers of filters, and consequently, parameters. In \Cref{fig:num_classes}, we plot the average of the logarithm of the EMC across different model sizes for various numbers of classes. We see that data with semantic labels becomes harder and harder to fit as the number of classes increases, and generalization becomes more challenging as the model has to encode more information about each sample in its weights. In contrast, randomly labeled data is easier to fit as the number of classes increases because the model is no longer forced to assign as many semantically different samples the same class label, which would be at odds with the model's inductive bias that prefers correct labels over random ones.

To compare different datasets while controlling for properties like number of classes, we convert several datasets into binary classification problems. This modification enables us to assess the impact of the number of classes on EMC and isolate the effects of input distribution. Our results (Appendix \Cref{fig:binary_datasets}) show that even though the EMC among image datasets increases in the binary classification setting over the original classification labels, tabular datasets consistently demonstrated higher EMC. Furthermore, significant differences persist among the different tabular datasets. These outcomes suggest that additional factors, perhaps intrinsic to the datasets themselves, contribute to EMC beyond the number of classes.

\subsection{Predicting generalization} Neural networks exhibit a marked preference for fitting semantically coherent labels over random ones, a tendency reflecting their inductive biases. This propensity, as depicted in \Cref{fig:complete3} (right), underscores a broader principle: a network's adeptness at fitting semantic labels compared to random ones often correlates with its generalization. Interestingly, this generalization enables certain architectures, like CNNs, to fit more samples than their parameter count might suggest, blurring the boundaries of over- and under-parameterization.

This observation bridges two seminal perspectives on model generalization. Traditional machine learning wisdom posits that high-capacity models tend to overfit, compromising their generalization on new data—a notion reflected in early generalization bounds, which are vacuous for neural networks \citep{vapnik1991principles, bartlett2002rademacher}. In contrast, PAC-Bayes theory proposes that a model's flexibility doesn't inherently impede generalization, provided its prior assigns disproportionate mass to the true labels compared to random ones, or in other words the model prefers correct labelings of the data to incorrect labelings \citep{dziugaite2017computing}. Our empirical findings relate these two theories, revealing an empirical relationship between a model's increased ability to fit correct labels over random ones and its generalization.

Specifically, we compute the EMC for various CNN and MLP configurations on both correctly and randomly labeled data. We measure the percent increase in EMC when models encounter semantic labels versus random ones, effectively gauging their practical capacity to fit data that aligns with natural label distributions.

The notable inverse correlation between this metric and the generalization gap (Pearson correlation coefficient of $-0.9281$ for CNNs and $-0.869$ for MLPs), as illustrated in \Cref{fig:complete3} (Right), not only confirms the theoretical underpinnings of generalization but also illuminates the practical implications of these theories.

\section{The Effect of Model Architecture on EMC}
After dissecting the influence of the data on flexibility, we shift focus to the impact of architecture. In this section, we examine how various architectural properties, including MLPs vs. CNNs vs. transformers, activation functions, and scaling strategies, contribute to flexibility.

\textbf{Architectural style and parameter efficiency.} There is an ongoing debate regarding the efficiency and generalization of CNNs and Vision Transformers (ViTs) \citep{d2021convit, patro2023efficiency, app13095521, goldblum2023battle}. In light of this debate, we put three neural network architecture paradigms to the test: (1) MLPs, (2) CNNs, and (3) ViTs.

Our findings, summarized in Appendix \Cref{fig:architectures}, reveal a consistent pattern: CNNs, characterized by hard-coded inductive biases like locality and translation equivariance, outperform both ViTs and MLPs in EMC (see \Cref{fig:complete1} for detailed EMC scaling laws across various parameter counts). This superiority persists across all model sizes when evaluated on semantically labeled data. As analyzed in the previous section, this trend could be misconstrued as purely a result of better generalization capabilities of CNNs over ViTs and, subsequently, MLPs.

To test this hypothesis, we examine the networks' flexibility on randomized data (\Cref{fig:complete1}). CNNs, which benefit strongly from data with a spatial structure, fit fewer samples when the spatial structure is broken via permutation. On the other hand, MLPs lack this preference for spatial structure, so their ability to fit data is unchanged. Replacing inputs with Gaussian noise increases both architectures' capacity. This trend might be explained by the fact that in high dimensions, noisy data lies far apart and is, therefore, easier to separate. Notably, CNNs can fit far more samples with semantic labels than with random inputs. In contrast, this trend is reversed for MLPs, again highlighting the superior generalization of CNNs on image classification.

However, even though random data impacts architectures differently, their hierarchy in terms of parameter efficiency remains unchanged. This intriguing consistency underscores that the enhanced parameter efficiency of CNNs is not just a byproduct of their generalization ability but is inherently rooted in their architectural design. This observation aligns with theoretical perspectives in approximation theory, which posit CNNs as more parameter-efficient compared to MLPs \citep{bao2014approximation}.

\textbf{Strategies for scaling network size.} The debate on how to best scale width and depth in neural networks has been ongoing. We now focus on how different scaling strategies affect a network's ability to fit data. Our examination, shown in \Cref{fig:scaling} and Appendix \Cref{fig:vit_scaling_laws}, explores EMC under various scaling configurations. For ResNets, these include increasing width (number of filters), increasing depth, or increasing both width and depth according to two scaling laws: EfficientNet \citep{tan2019efficientnet} and ResNet-RS \citep{bello2021revisiting}. EfficientNet uses a balanced approach, scaling depth, width, and resolution simultaneously with fixed coefficients. ResNet-RS adapts scaling based on model size, training duration, and dataset size. For scaling ViTs, we use the SViT approach \citep{zhai2022scaling}, SoViT \citep{alabdulmohsin2023getting}, and also try scaling the number of encoder blocks (depth) and the dimensionality of patch embeddings and self-attention (width) separately.

Our analysis reveals that, although not initially crafted for optimizing capacity, specially designed scaling laws perform well in this respect. Furthermore, consistent with earlier theoretical analyses \citep{eldan2016power}, our findings affirm that scaling depth is more parameter-efficient than scaling width. These parameter-efficiency comparisons also hold on randomly labeled data, indicating that they are not an artifact of generalization.

\begin{figure}[t!]
    \centering
    \begin{subfigure}{0.46\textwidth}
        \centering
        \includegraphics[width=1\linewidth]{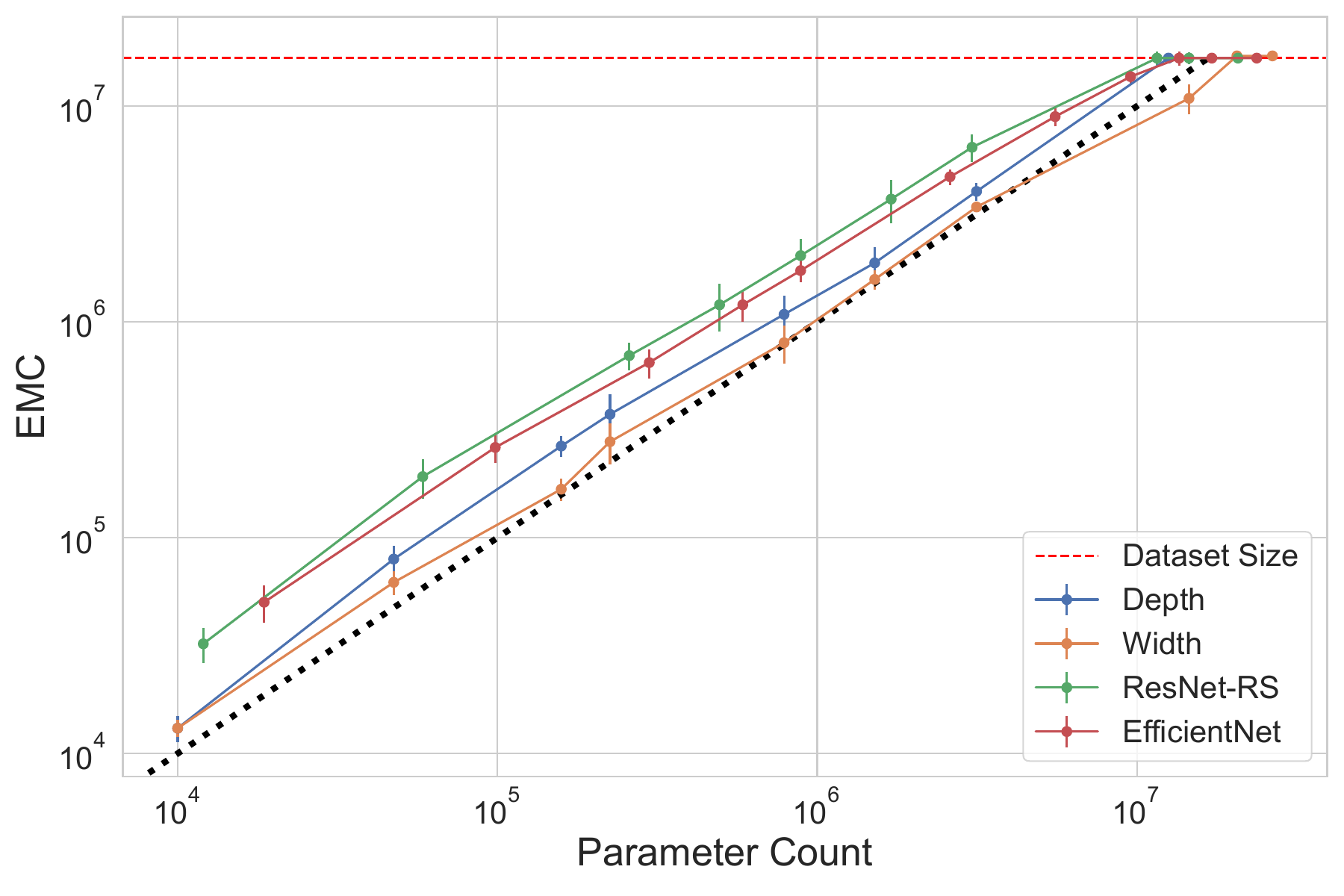}
    \caption{\textbf{ResNet-RS is the most efficient among scaling strategies we test.}}  
            \label{fig:scaling}
    \end{subfigure}
    \hfill
    \begin{subfigure}{0.53\textwidth}
       \centering
        \centering
        \includegraphics[width=1\linewidth]{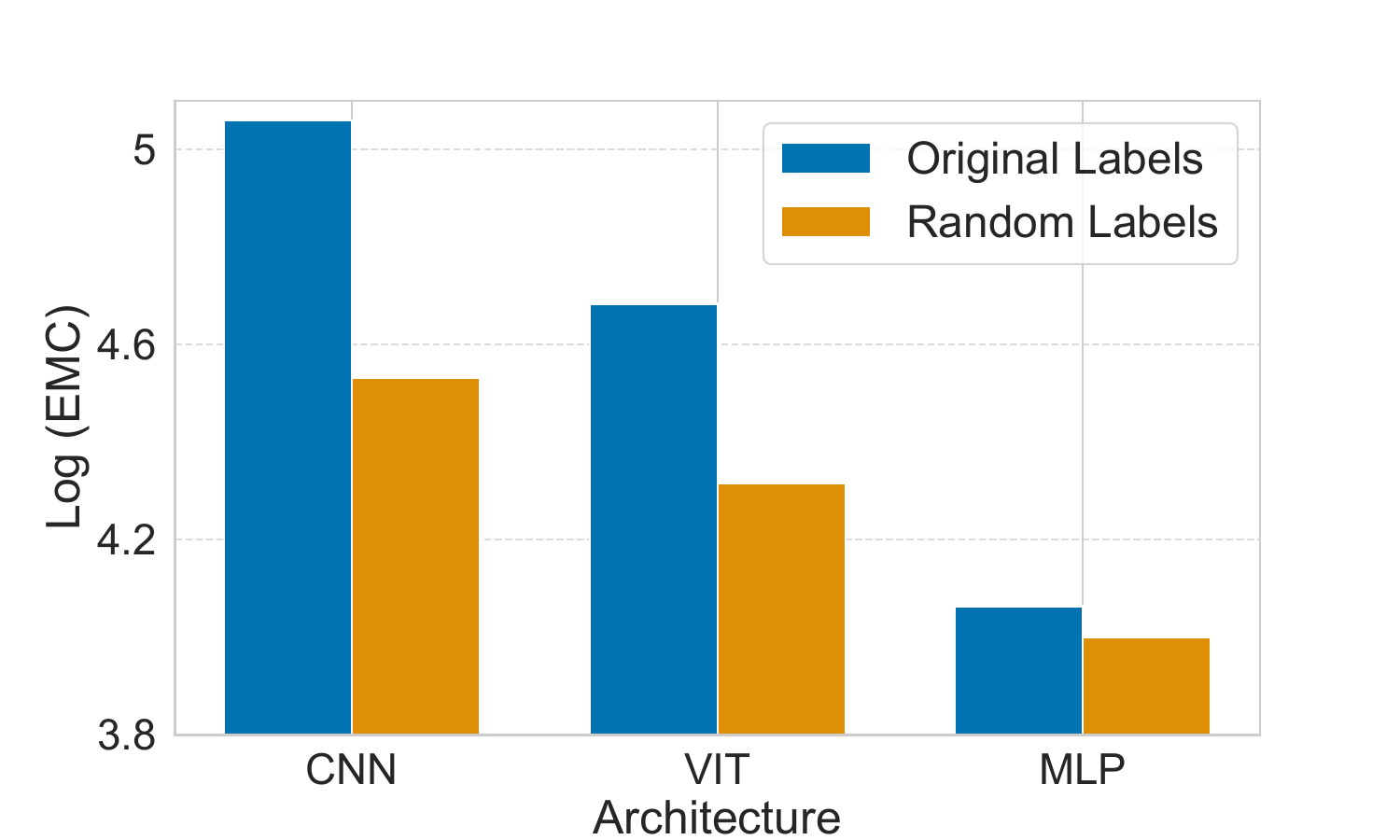}
    \caption{\textbf{CNNs are far more parameter-efficient, even on randomly labeled data.} }  
    \label{fig:architectures}
    \end{subfigure}
    \caption{\textbf{The effect of the scaling strategy and the architecture on the EMC} . \textbf{(a)}  Scaling laws for the EMC as a function of parameters counts for CNN. \textbf{(b)} Average logarithm of EMC across parameter counts for different architectures using original and random labels.   On ImageNet-20MS. Error bars represent one standard error over 5 trials.  }
    \label{fig:random232}
\end{figure}

\textbf{Activation functions.} Nonlinear activation functions are crucial for neural network capacity because without them, neural networks are just large factorized linear models. In this subsection, we examine the effect of the activation functions on capacity, contrasting them with linear models.

Detailed in Appendix \Cref{fig:activation_func_random}, our findings show that ReLU functions significantly enhance capacity. Though initially integrated to mitigate vanishing and exploding gradients, ReLU also boosts the network's data-fitting ability, likely by improving generalization. In contrast, tanh and identity functions, while nonlinear, do not achieve similar effects, even though we are able to find minima using these activations functions too.  We note the latter fact to ensure that ReLUs are not only boosting network capacity by making it easier to find minima.

\section{The Role of Optimization in Fitting Data}

The choice of optimization technique and regularization strategy is crucial in neural network training. This choice affects not only training convergence but also the nature of the solutions found. This section explores the role different optimization and regularization techniques play in a network's flexibility.

\textbf{Comparing optimizers.} We explore the influence of various optimizers, including SGD, full-batch Gradient Descent, Adam \citep{kingma2014adam}, AdamW \citep{loshchilov2018decoupled}, and Shampoo \citep{gupta2018shampoo}.

Whereas previous works suggest that SGD has a strong flatness-seeking regularization effect \citep{geiping2021stochastic},
we find in \Cref{fig:optimizers} that SGD also enables fitting more data than full-batch (non-stochastic) training, fitting a comparable volume of data as the high-powered Shampoo.  This experiment, namely the variety of EMC measurements across optimizers, demonstrates that optimizers differ not only in the rate at which they converge but also in the types of minima they find.  Repeating this experiment with random labels  shows that the higher EMC of SGD and Shampoo evaporates, indicating that their greater ability to fit data may be related to their superior generalization.

\textbf{Regularizers.} Classical machine learning systems employed regularizers designed to reduce capacity.  For example, ridge regression applies a penalty on the parameter norm, improving performance of overparameterized linear models \citep{hoerl1970ridge}.  Similarly XGBoost penalizes the sum of squared leaf weights to prevent overfitting \citep{chen2016xgboost}. Modern deep learning pipelines use various regularization techniques to improve generalization. We now examine if these regularizers also reduce the model's capacity to fit data. We previously found that stochastic training, which enhances generalization and provides implicit regularization, actually increases EMC.

In Appendix \Cref{fig:reg}, we compute the EMC of a CNN trained on ImageNet-20MS using Sharpness-Aware Minimization (SAM) \citep{foret2020sharpness}, weight decay, and label smoothing \citep{muller2019does}. Weight decay and label smoothing limit capacity, but SAM improves generalization without reducing capacity, even on randomly labeled data. Label smoothing modifies the loss function, so a model trained with the smoothed objective may not find minima of the original non-smoothed loss. In contrast, SAM does not change the loss function itself but finds different types of minima than SGD, which generalize better at no capacity cost.

\section{Reparameterization for Increased Parameter Efficiency}
\label{sec:reparameterization}

In previous sections, we examined how different pipeline components affect parameter efficiency. We found that neural networks often fit fewer samples than they have parameters. To close this gap, we adopt two reparameterization methods - \emph{subspace training} \citep{lotfi2022pac}, which was designed for compression, whereby we take a CNN's parameter vector and randomly project it into a lower-dimensional subspace, training in the lower dimensional space.  We also try a quantization experiment where we train the CNN in 8-bit precision instead of the standard 32-bit precision but with four times as many parameters. This quantization deviates from our other parameter-count studies. Here, we count a model with $4 \times n$ 8-bit parameters with a model containing $n$ 32-bit parameters as they are specified by the same number of bits.

Our empirical observations, detailed in Appendix \Cref{fig:projection}, underscore the effectiveness of these strategies. Subspace training, in particular, significantly improves parameter efficiency for both semantic and random labels. This highlights that neural networks, in their standard form, are wasteful of parameters. The quantization experiment further supports this finding, revealing that even with reduced precision, networks can achieve comparable levels of flexibility. Interestingly, we also see here that an 8-bit quantized model can fit a quarter times as many randomly labeled samples as it has parameters, closing the same gap on a per-bit basis.

\section{Discussion}

Our findings show that parameter counting alone is not a useful tool for determining the number of samples a neural network can fit, or the boundary between underparameterization and overparameterization.  Instead, many factors contribute to the effective model complexity, including virtually all components of a training routine as well as the data itself.  Moreover, we must re-evaluate our understanding of why these components work.  We saw that architectural components like ReLU activation functions may solve additional problems that they weren’t designed for, and stochastic optimization, for example, actually finds minima where we fit more training samples, contrasting with conventional views of implicit regularization. Finally, our results suggest neural networks are often parameter-wasteful, and new parameterizations might improve efficiency.

\subsection*{Acknowledgements}
This work is supported by NSF CAREER IIS-2145492,
NSF CDS\&E-MSS 2134216, NSF HDR-2118310, BigHat Biosciences, Capital One, and an Amazon Research Award.

\clearpage

\bibliography{references} 
\clearpage
\appendix
\section{Appendix}
\subsection{Additional Results}

Here, we present figures that include additional datasets and labelings, as well as detailed results across all parameter counts, rather than just the aggregated averages shown in the main body. In the main paper, for the ViT scaling laws, we followed the scaling approach proposed by \cite{zhai2022scaling} (SVIT), which advocates for simultaneously and uniformly scaling all aspects—depth, width, MLP width, and patch size. Additionally, we employed both SoViT, as per \cite{alabdulmohsin2023getting}, and approaches where the number of encoder blocks (depth) and the dimensionality of patch embeddings and self-attention (width) in the ViT are scaled separately. \cref{fig:vit_scaling_laws} in the Appendix demonstrates that scaling each dimension independently can lead to suboptimal results, aligning with our observations from the EfficientNet experiments. Furthermore, it shows that SoViT yields results that are slightly different from those obtained using the laws from \cite{zhai2022scaling}.

\begin{figure}[H]
\centering
\includegraphics[width=0.95\linewidth]{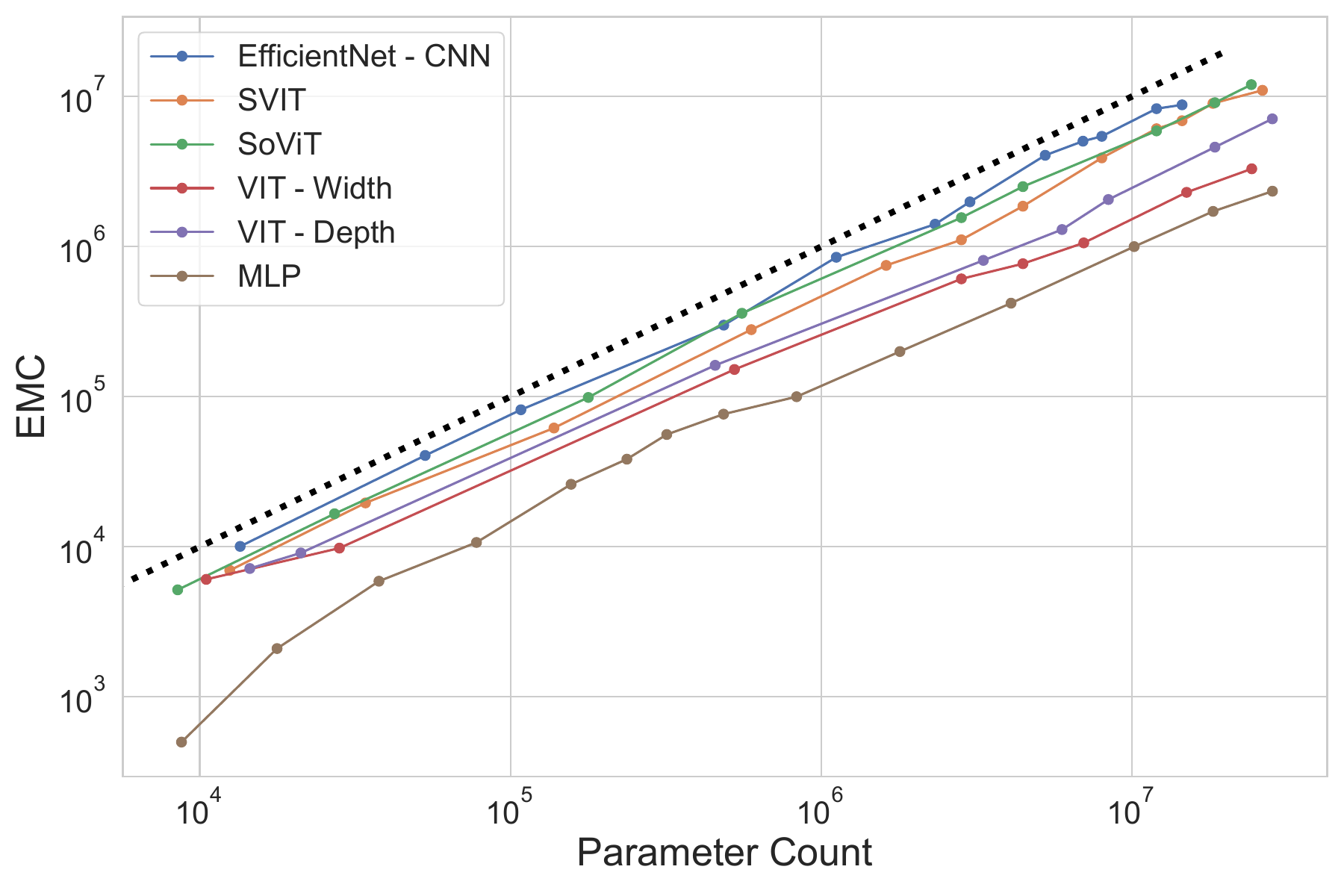}
\caption{\textbf{Scaling laws -} EMC as a function of the number of parameters for randomly labeled ImageNet-20MS for VIT}
\label{fig:vit_scaling_laws}
\end{figure}

\begin{figure}[H]
\centering
\includegraphics[width=0.95\linewidth]{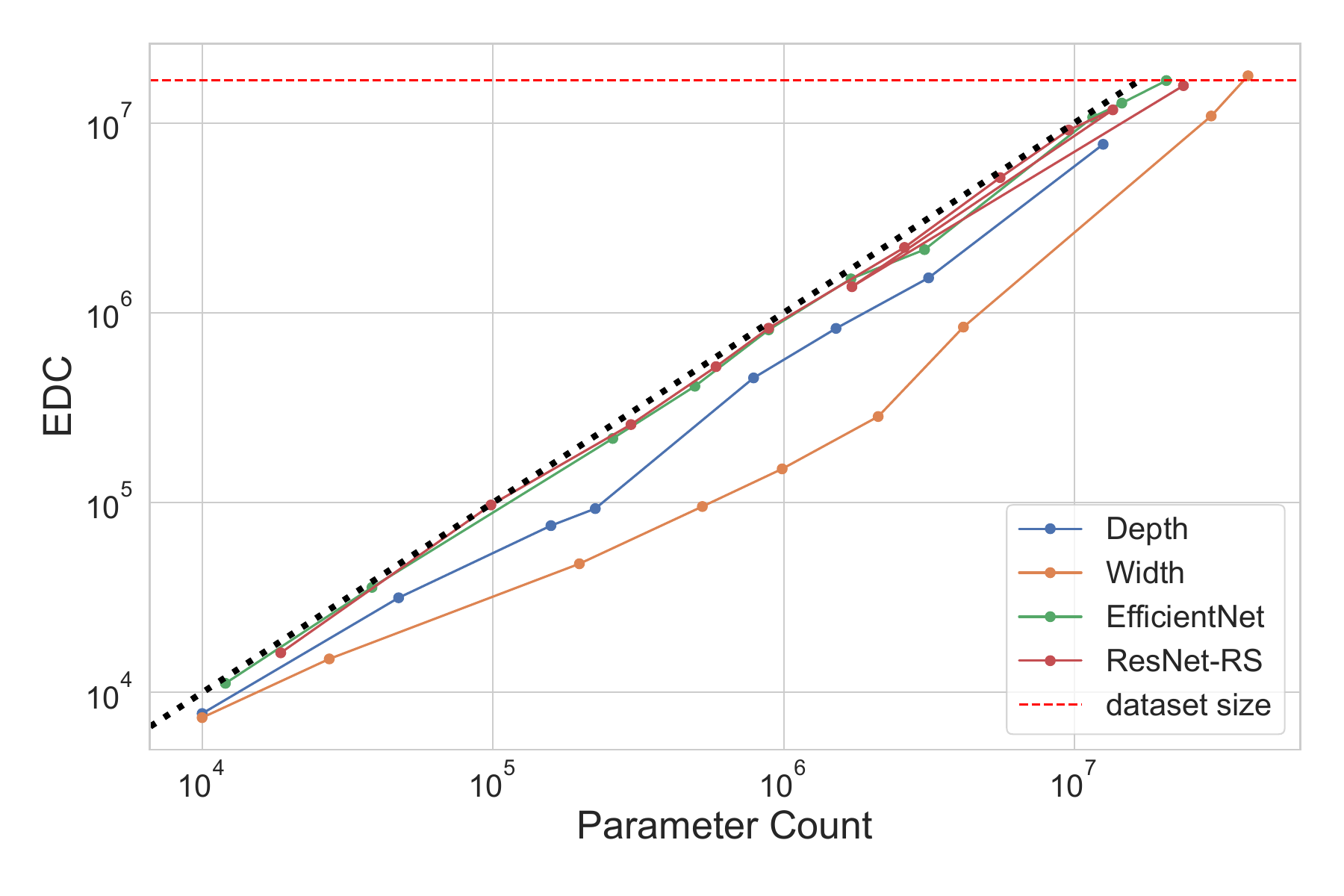}
\caption{\textbf{Scaling laws -} EMC as a function of the number of parameters for randomly labeled ImageNet-20MS.}
\label{fig:num_classes2}
\end{figure}

\begin{figure}[H]
\centering
\includegraphics[width=0.95\linewidth]{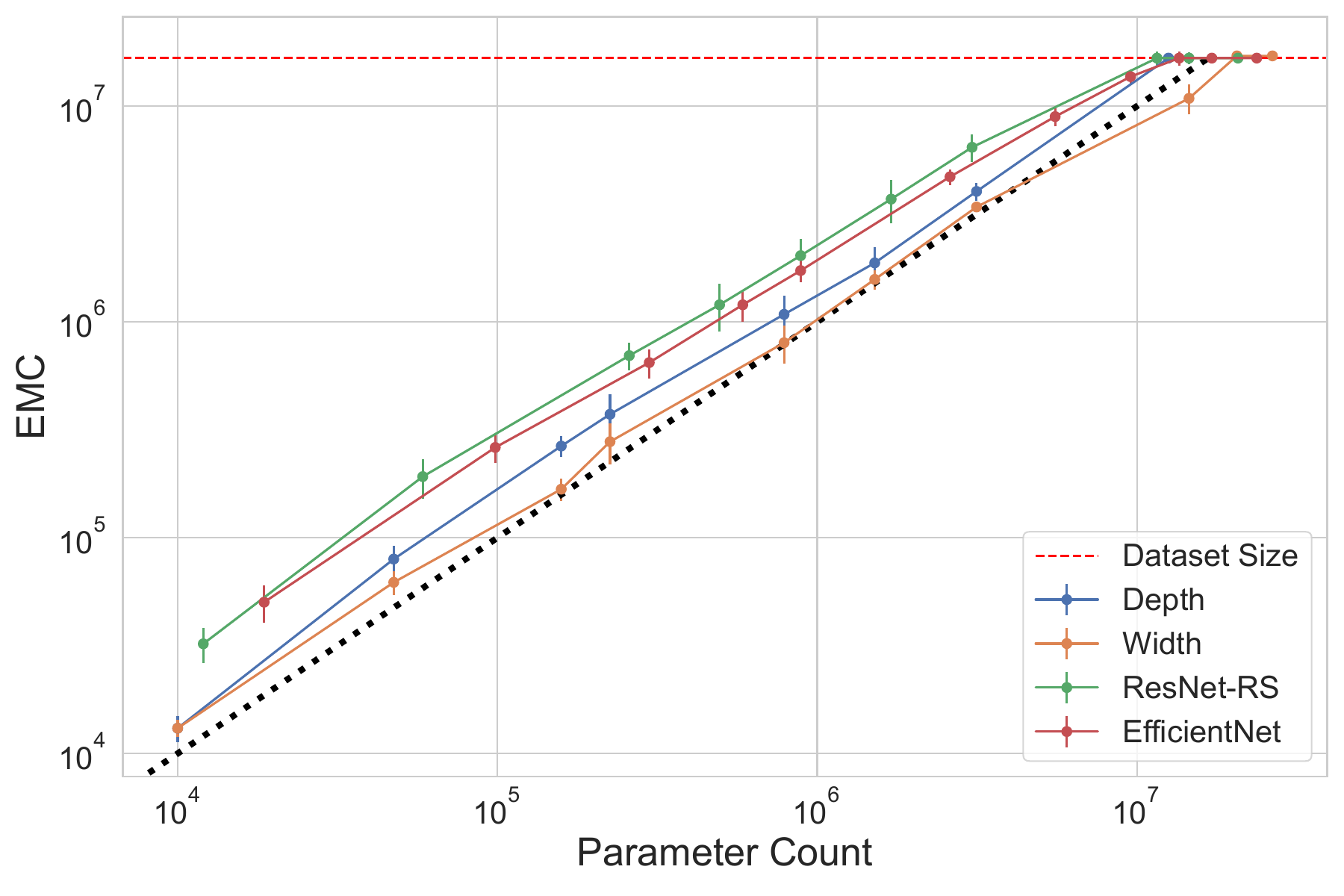}
\caption{\textbf{Scaling laws -} EMC as a function of the number of parameters for a CNN on ImageNet-20MS with original labels.}
\label{fig:num_classes3}
\end{figure}

\begin{figure}[H]
\centering
\includegraphics[width=0.95\linewidth]{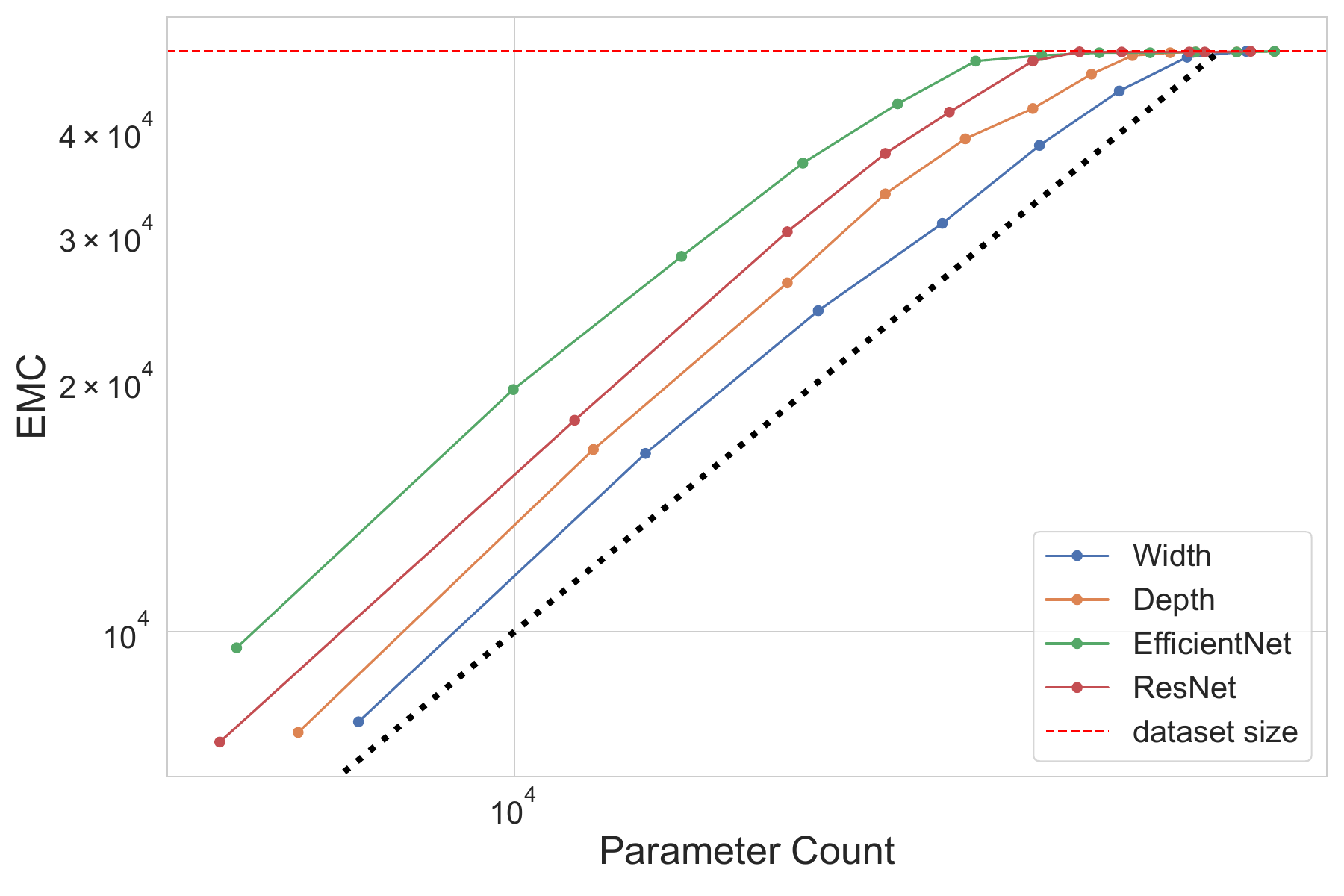}
\caption{\textbf{Scaling laws -} EMC as a function of the number of parameters for a CNN on CIFAR-10 with original labels.}
\label{fig:num_classes4}
\end{figure}

\begin{figure}[H]
\centering
\includegraphics[width=0.95\linewidth]{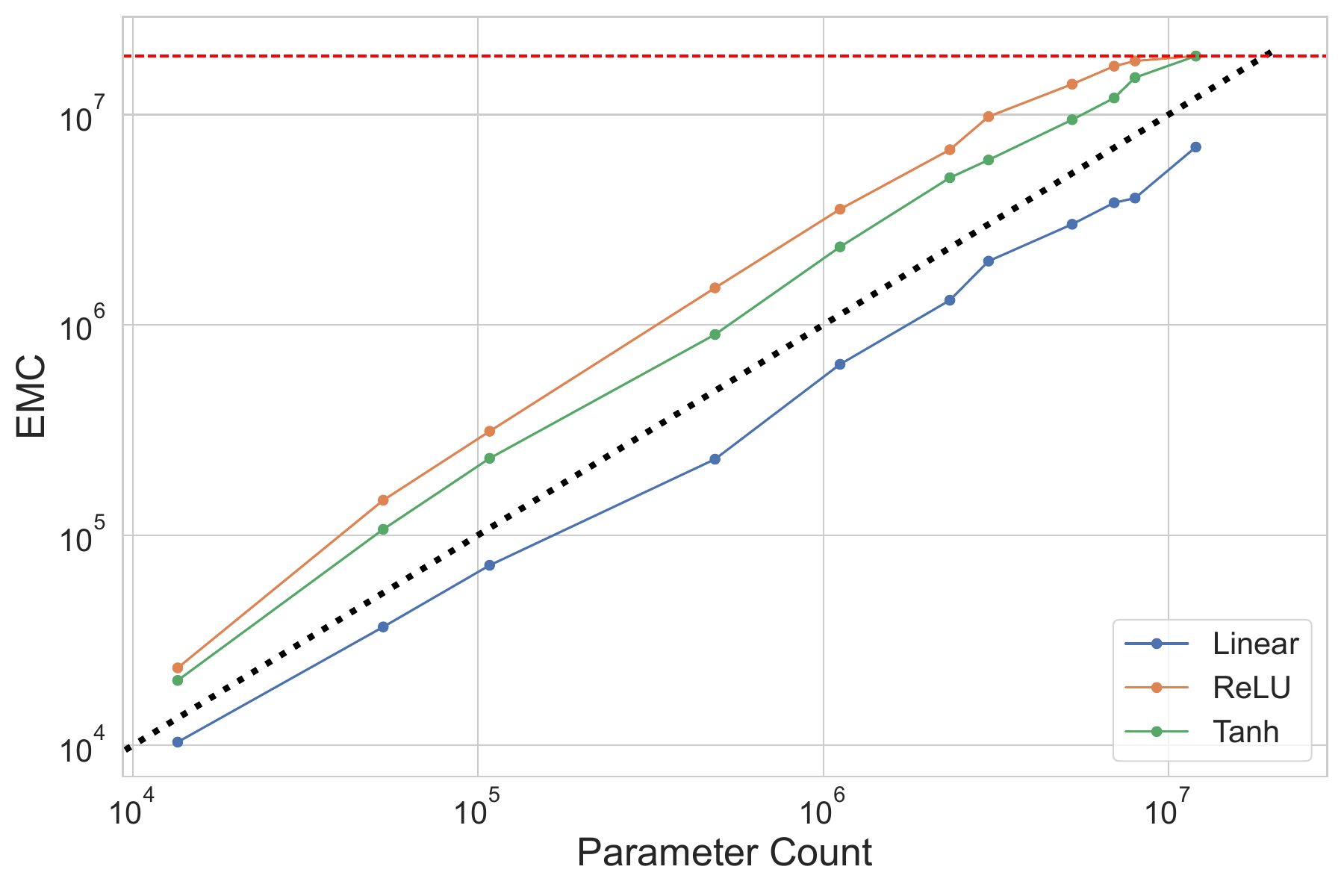}
\caption{\textbf{EMC as a function of the number of parameters across different activation functions} using CNNs on ImageNet-20MS with original labels.}
\label{fig:activation_func}
\end{figure}

\begin{figure}[H]
\centering
\includegraphics[width=0.95\linewidth]{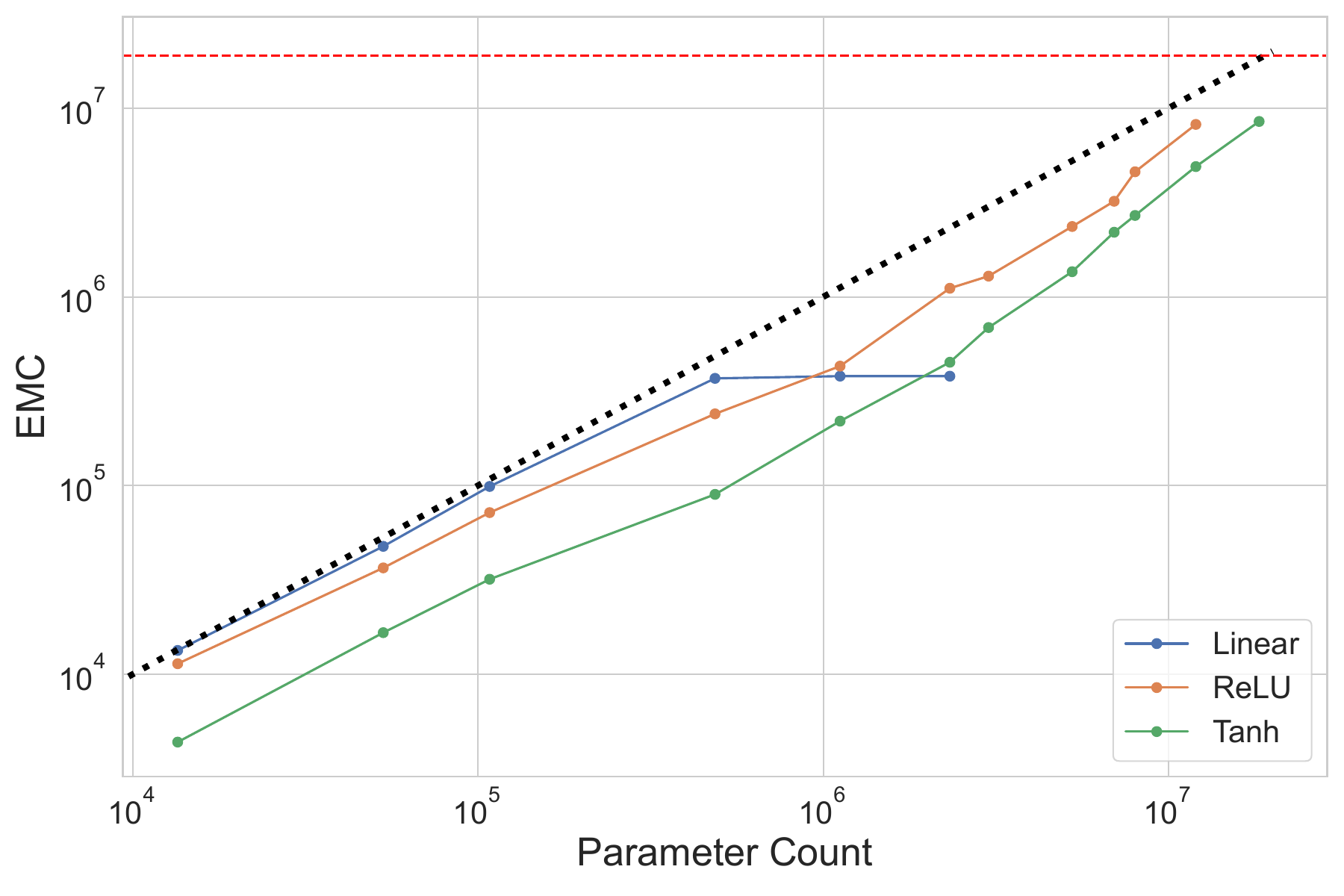}
\caption{\textbf{EMC as a function of the number of parameters across different activation functions} using CNNs and ImageNet-20MS with random labels.}
\label{fig:activation_func2}
\end{figure}

\begin{figure}[H]
\centering
\includegraphics[width=0.95\linewidth]{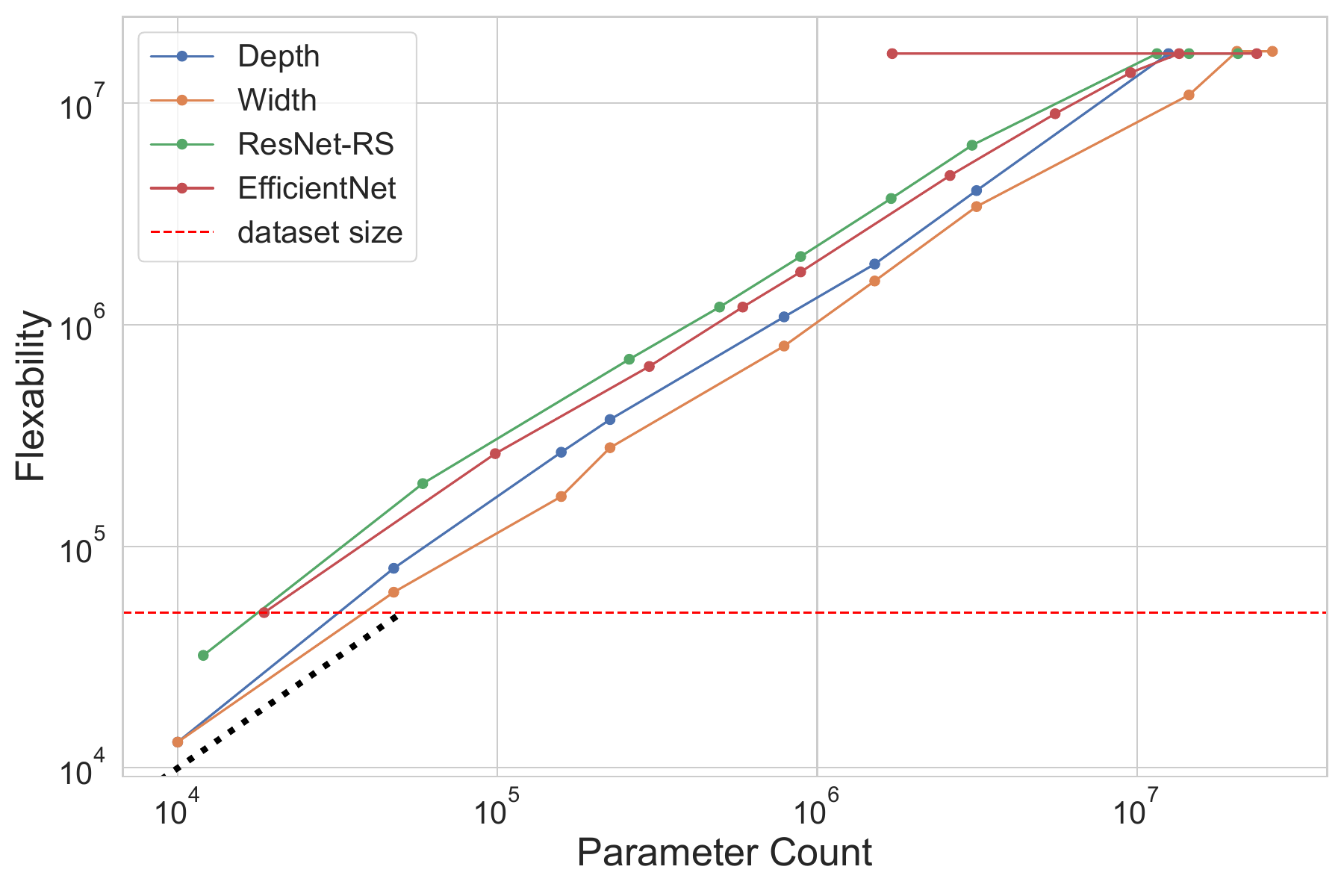}
\caption{\textbf{SGD and Shampoo fit more training data} - EMC across different optimizers using CNNs on CIFAR-10.}
\label{fig:optimizers3}
\end{figure}

\begin{figure}[H]
\centering
\includegraphics[width=0.95\linewidth]{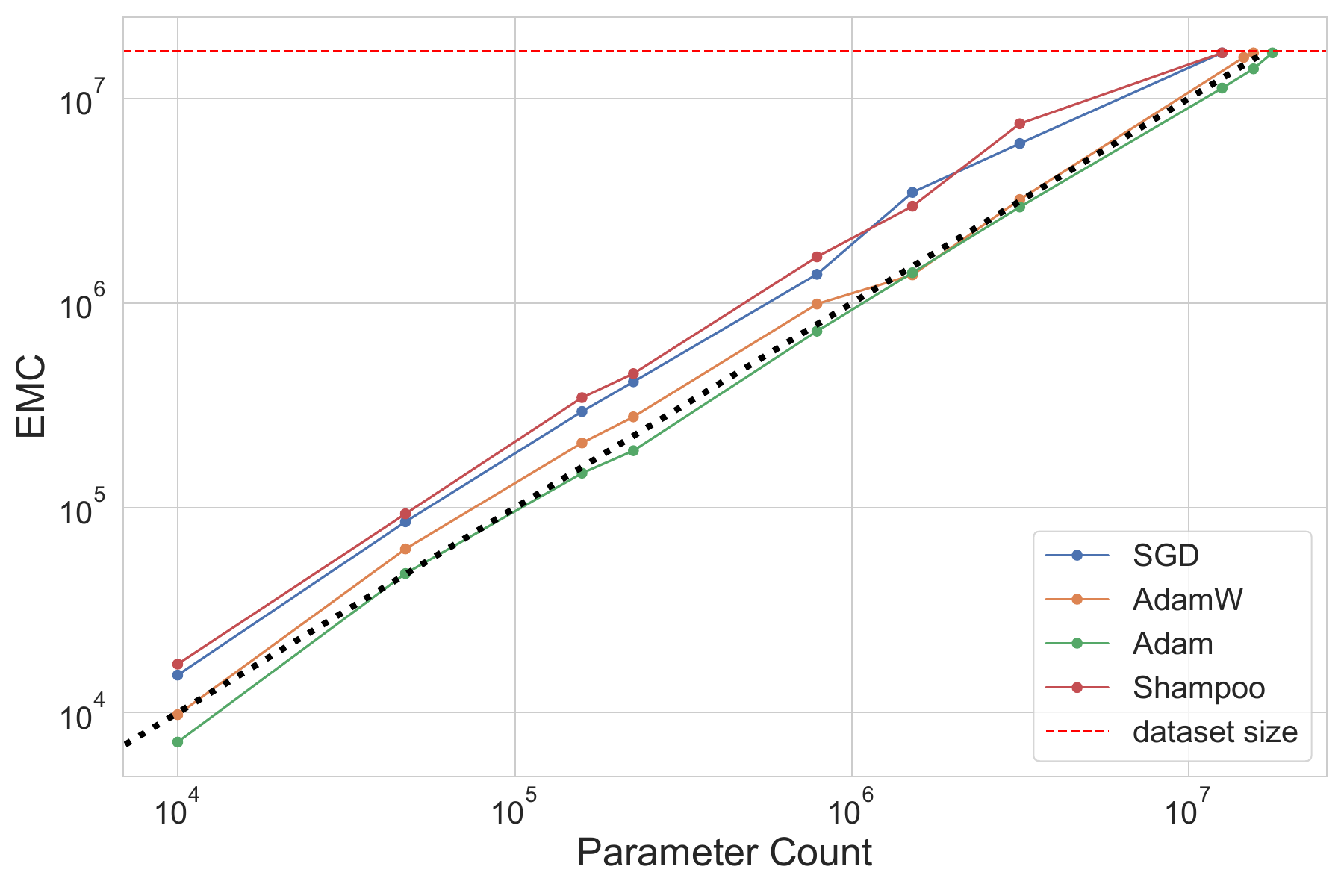}
\caption{\textbf{EMC as a function of the number of parameters across different optimizers} with CNNs on ImageNet-20MS with original labels.}
\label{fig:scaling2}
\end{figure}

\begin{figure}[H]
\centering
\includegraphics[width=0.95\linewidth]{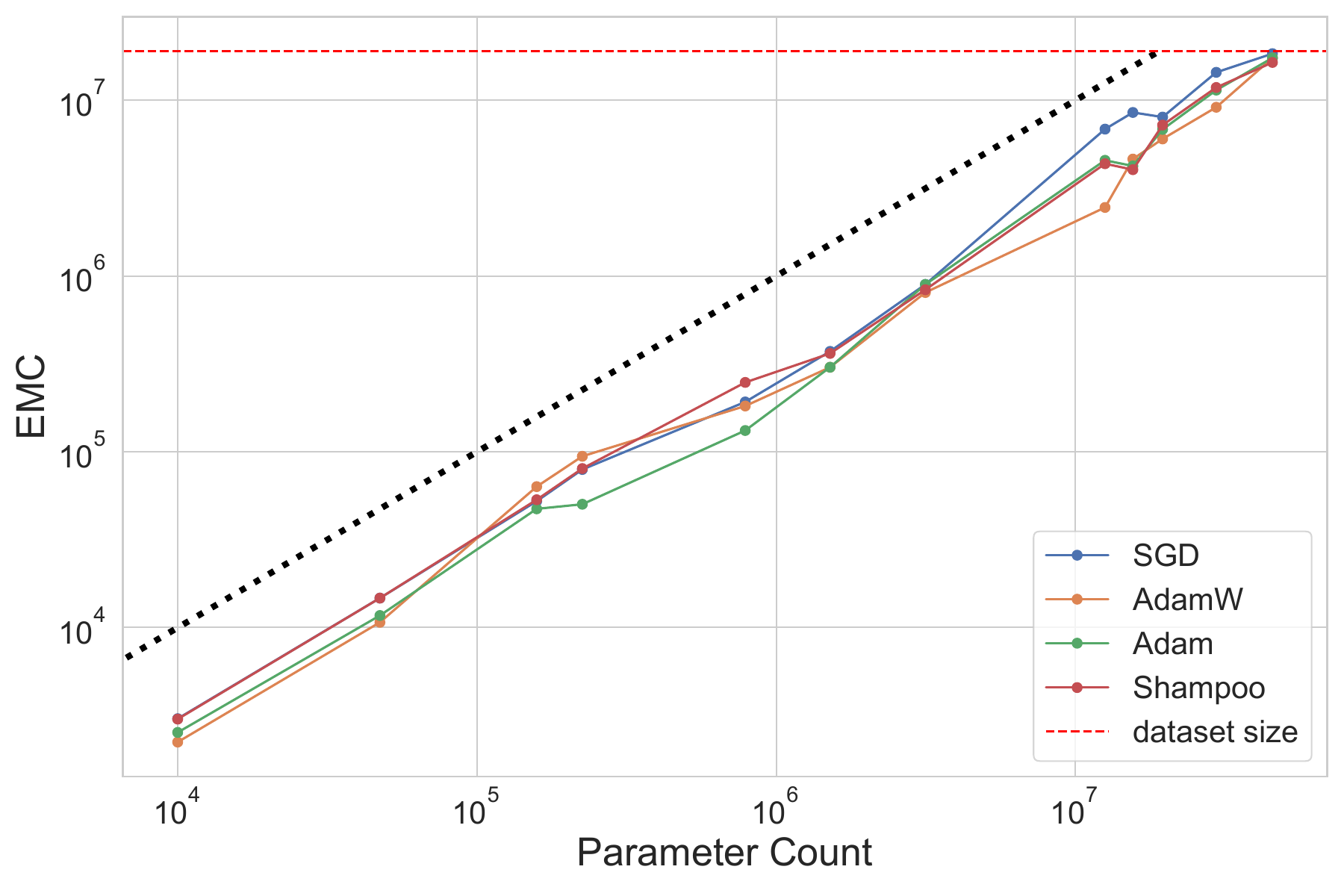}
\caption{\textbf{EMC as a function of the number of parameters across different optimizers} with CNNs on ImageNet-20MS with random labels.}
\label{fig:scaling3}
\end{figure}

\begin{figure}[H]
\centering
\includegraphics[width=0.95\linewidth]{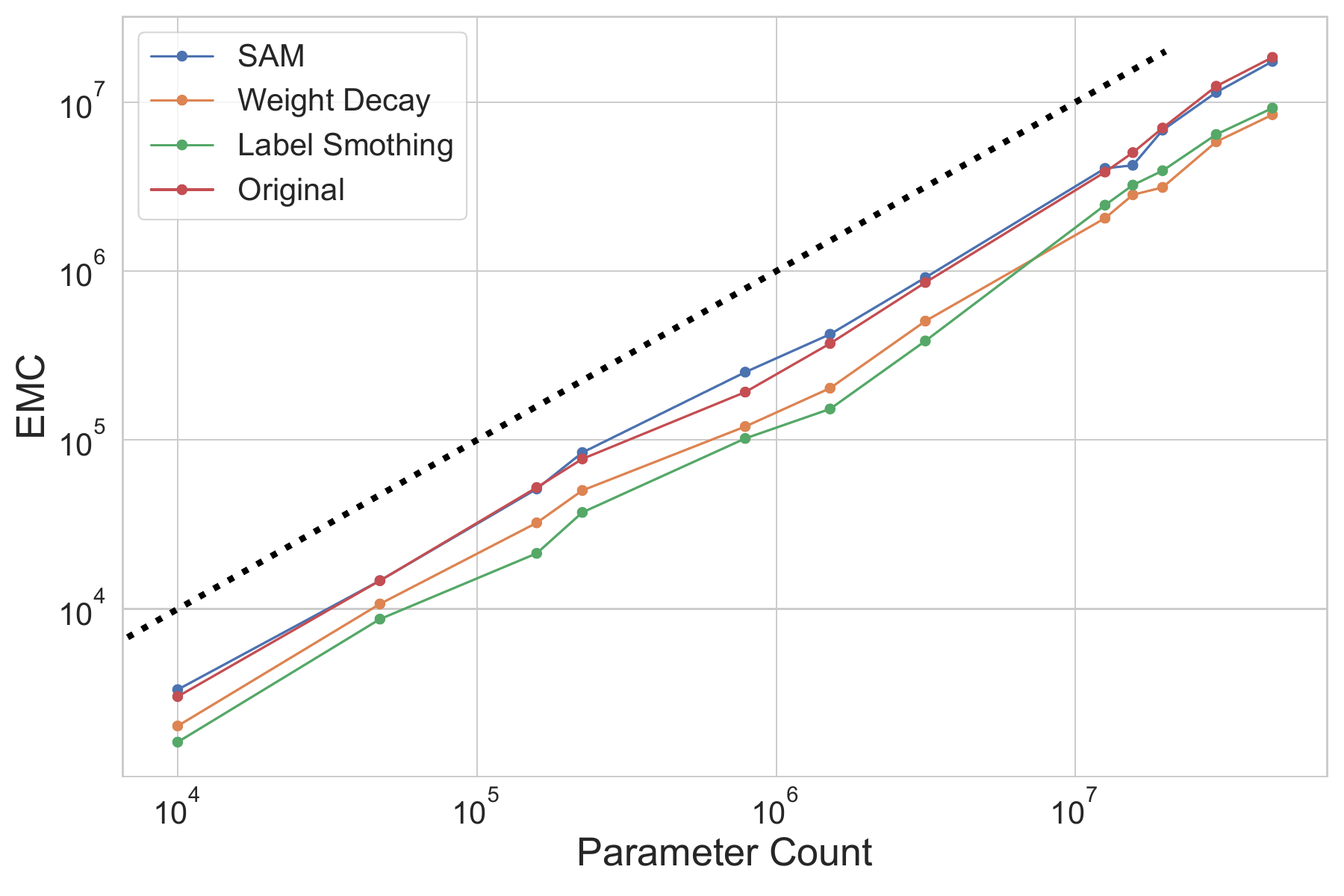}
\caption{\textbf{EMC as a function of the number of parameters across different regularizers} on ImageNet-20MS with random labels.}
\label{fig:reg2}
\end{figure}

\begin{figure}[H]
\centering
\includegraphics[width=0.95\linewidth]{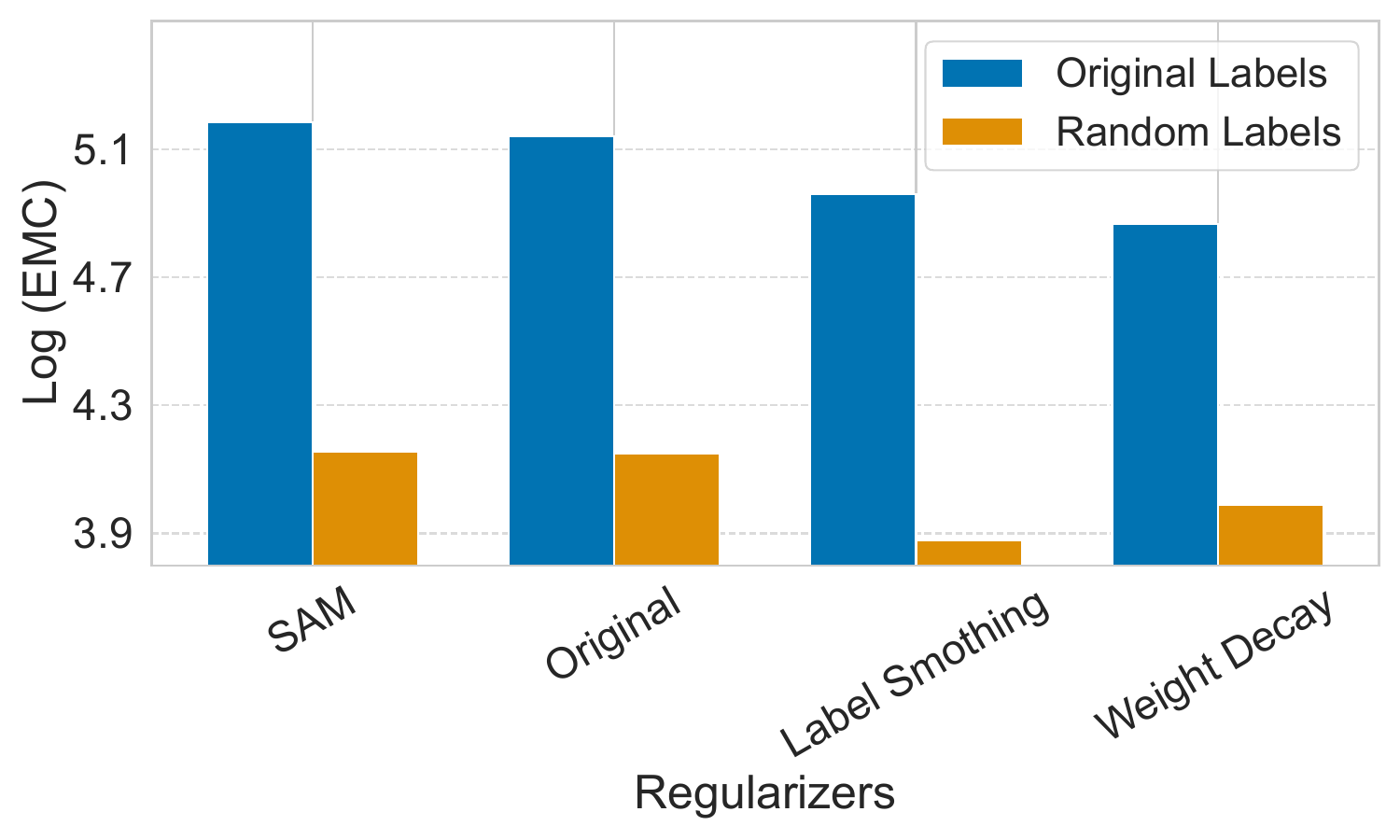}
\caption{\textbf{SAM has better generalization at no capacity cost -} Average logarithm of EMC over different model sizes for SAM, weight decay, and label smoothing using CNNs on ImageNet-20MS.}
\label{fig:reg}
\end{figure}

\begin{figure}[H]
\centering
\includegraphics[width=1\linewidth]{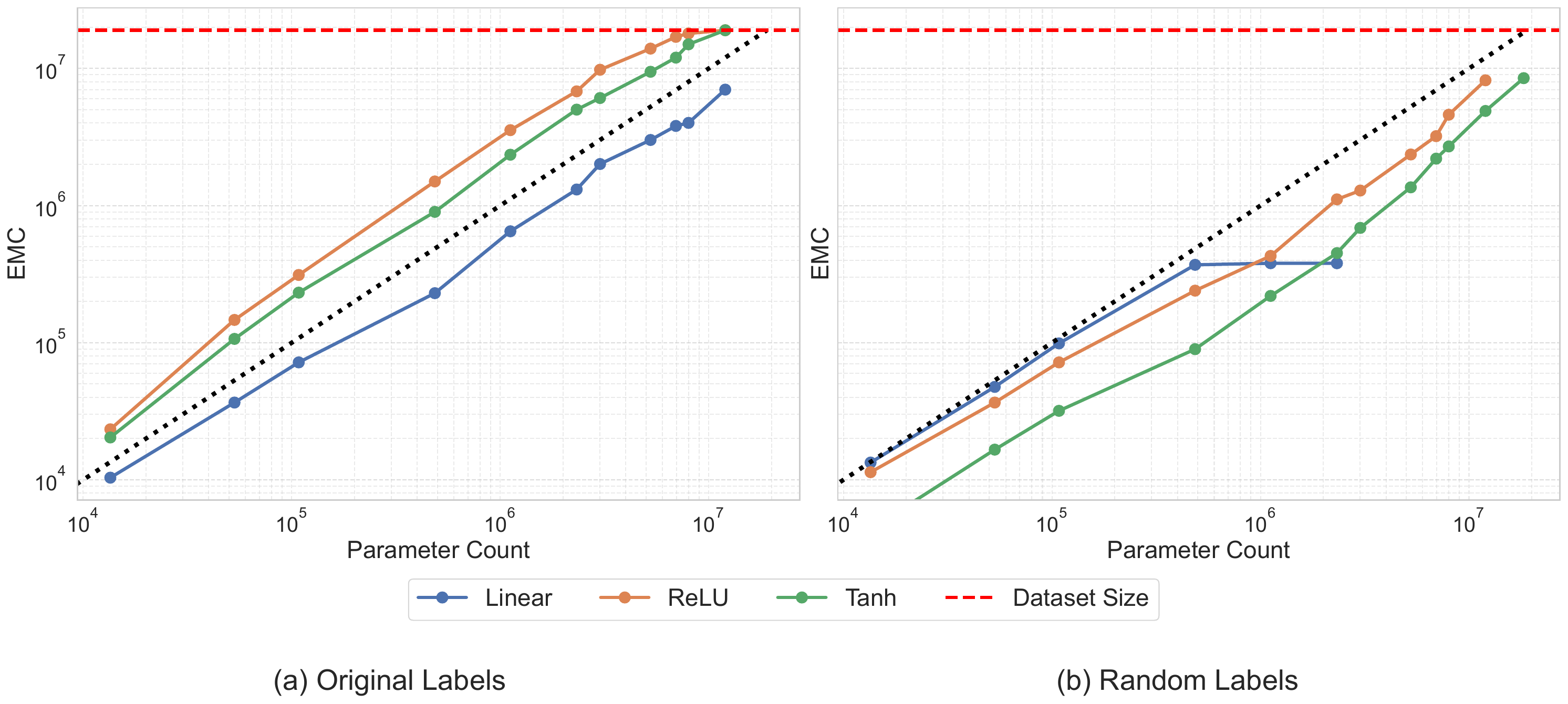}
\caption{\textbf{ReLU networks exhibit higher flexibility.} EMC as a function of the number of parameters across different activation functions for original labels \textbf{(left)} and for random ones \textbf{(right)} on ImageNet-20MS.}
\label{fig:activation_func_random}
\end{figure}

\begin{figure}[H]
\centering
\includegraphics[width=0.95\linewidth]{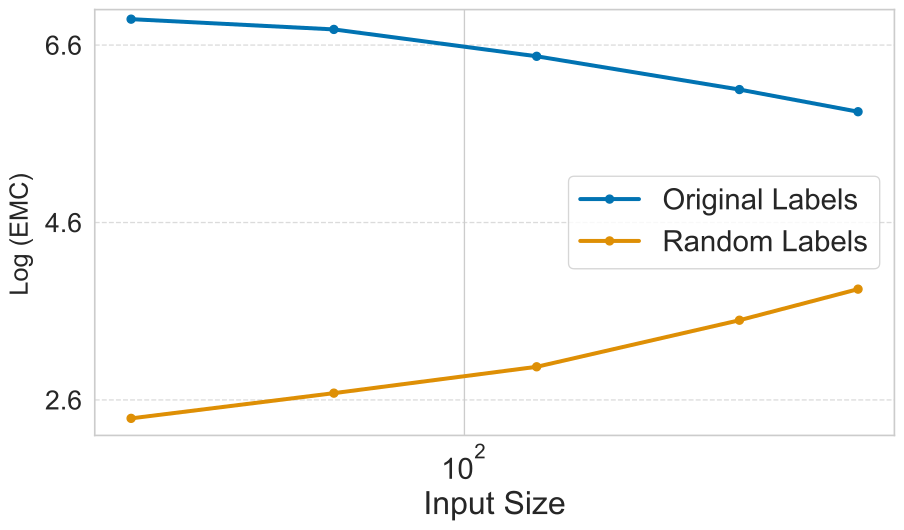}
\caption{\textbf{High-dimensional data is harder to fit.} Average logarithm of EMC across different model sizes for original and random labels varying input sizes for CNN architectures on CIFAR-100.}
\label{fig:input_size_flexibility}
\end{figure}

\begin{figure}[H]
    \centering
    \begin{subfigure}{0.48\textwidth}
        \centering
        \includegraphics[width=1\linewidth]{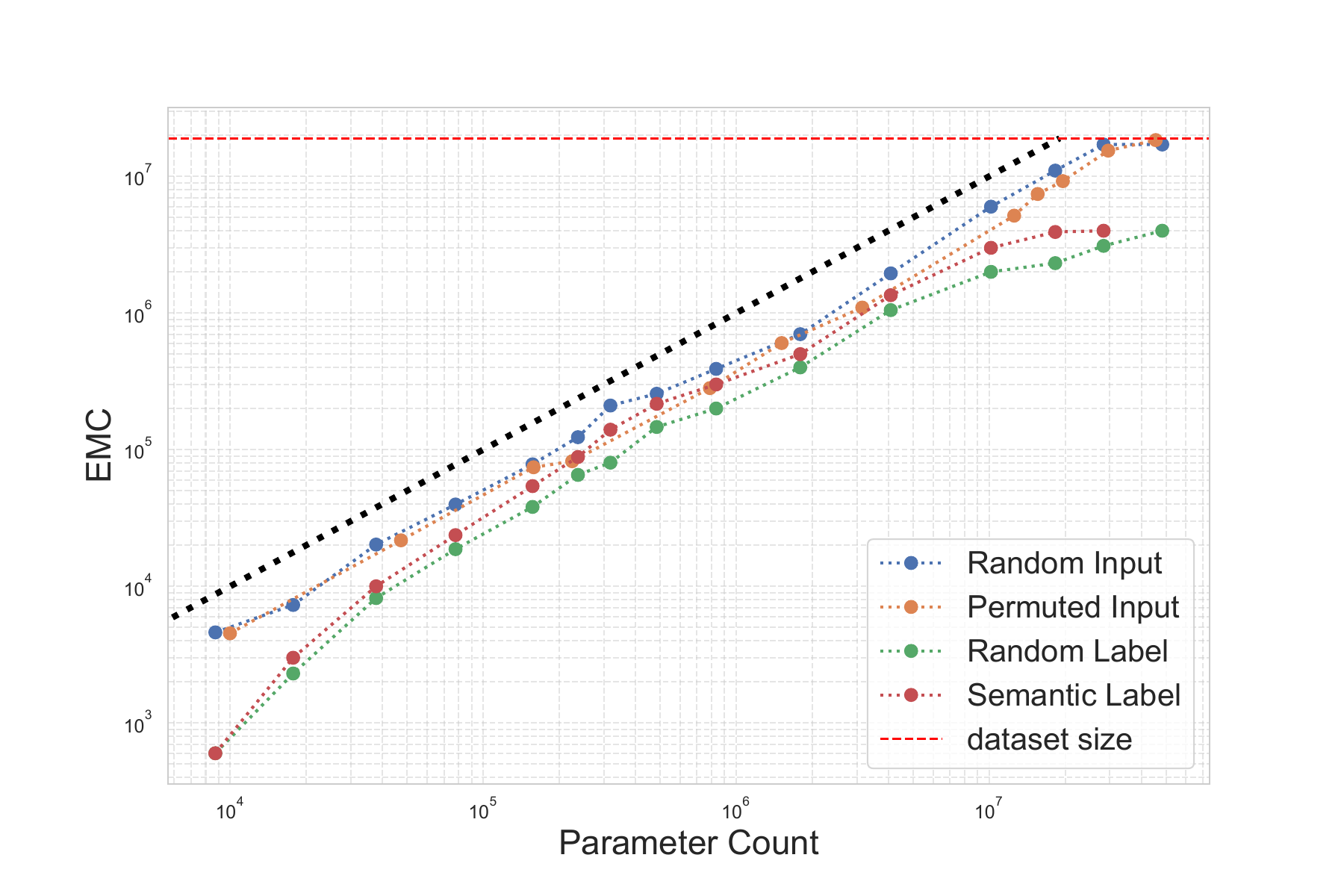}
        \caption{MLP}
        \label{fig:suba4}
    \end{subfigure}
    \hfill
    \begin{subfigure}{0.48\textwidth}
        \centering
        \includegraphics[width=1\linewidth]{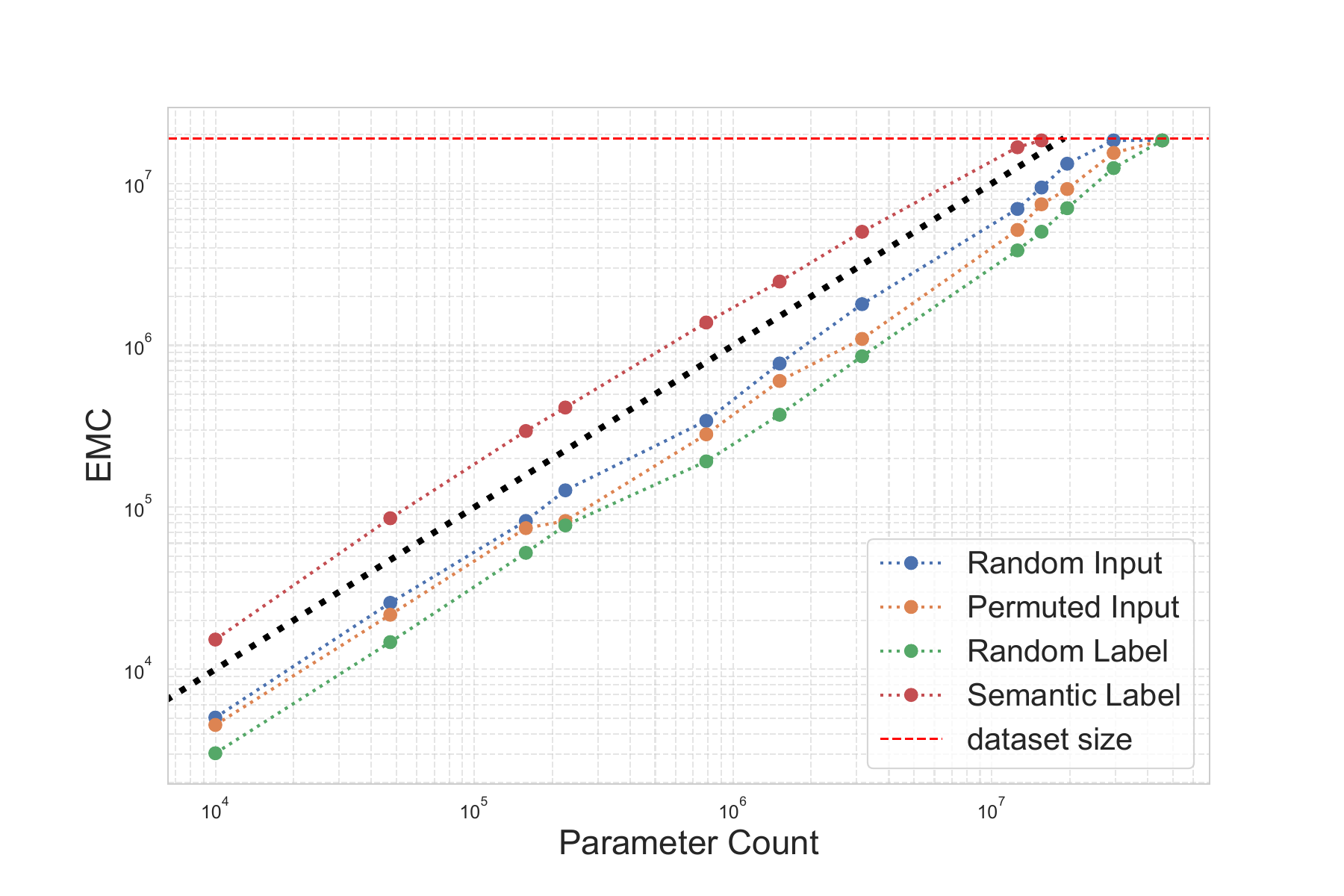}
        \caption{CNN}
        \label{fig:subb3}
    \end{subfigure}
    \caption{\textbf{Generalization boosts EMC -} EMC as a function of the number of parameters for semantic labels vs. random input and labels using MLP and CNN architectures on ImageNet-20MS.}  
    \label{fig:complete1}
\end{figure}

\begin{figure}[H]
\centering
\includegraphics[width=0.95\linewidth]{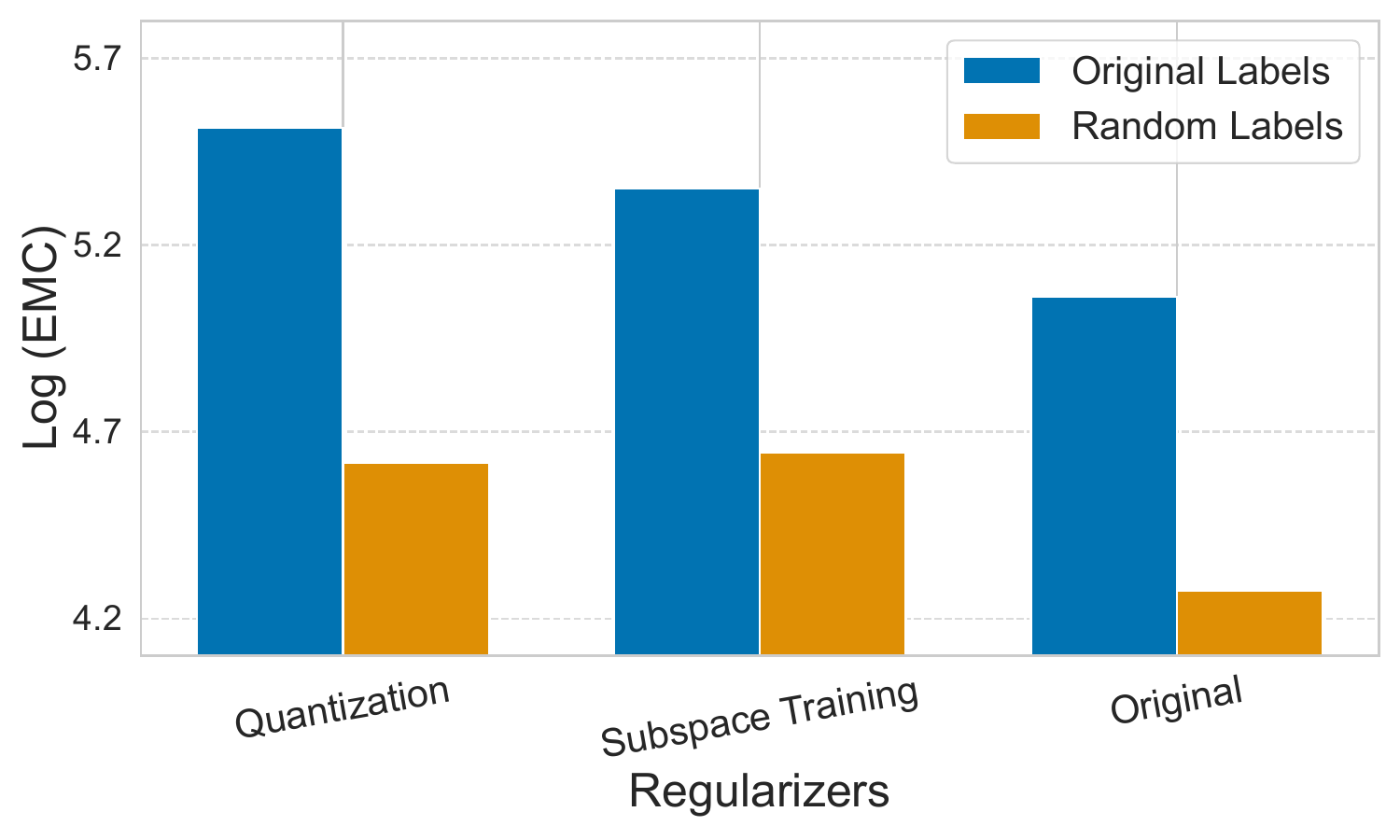}
\caption{\textbf{Compression improves network efficiency - } Average logarithm of EMC over different model sizes and compression methods. CNNs on ImageNet-20MS.}
\label{fig:projection}
\end{figure}

\begin{figure}[H]
    \centering
   \begin{subfigure}{0.48\textwidth}
        \centering
        \includegraphics[width=1\linewidth]{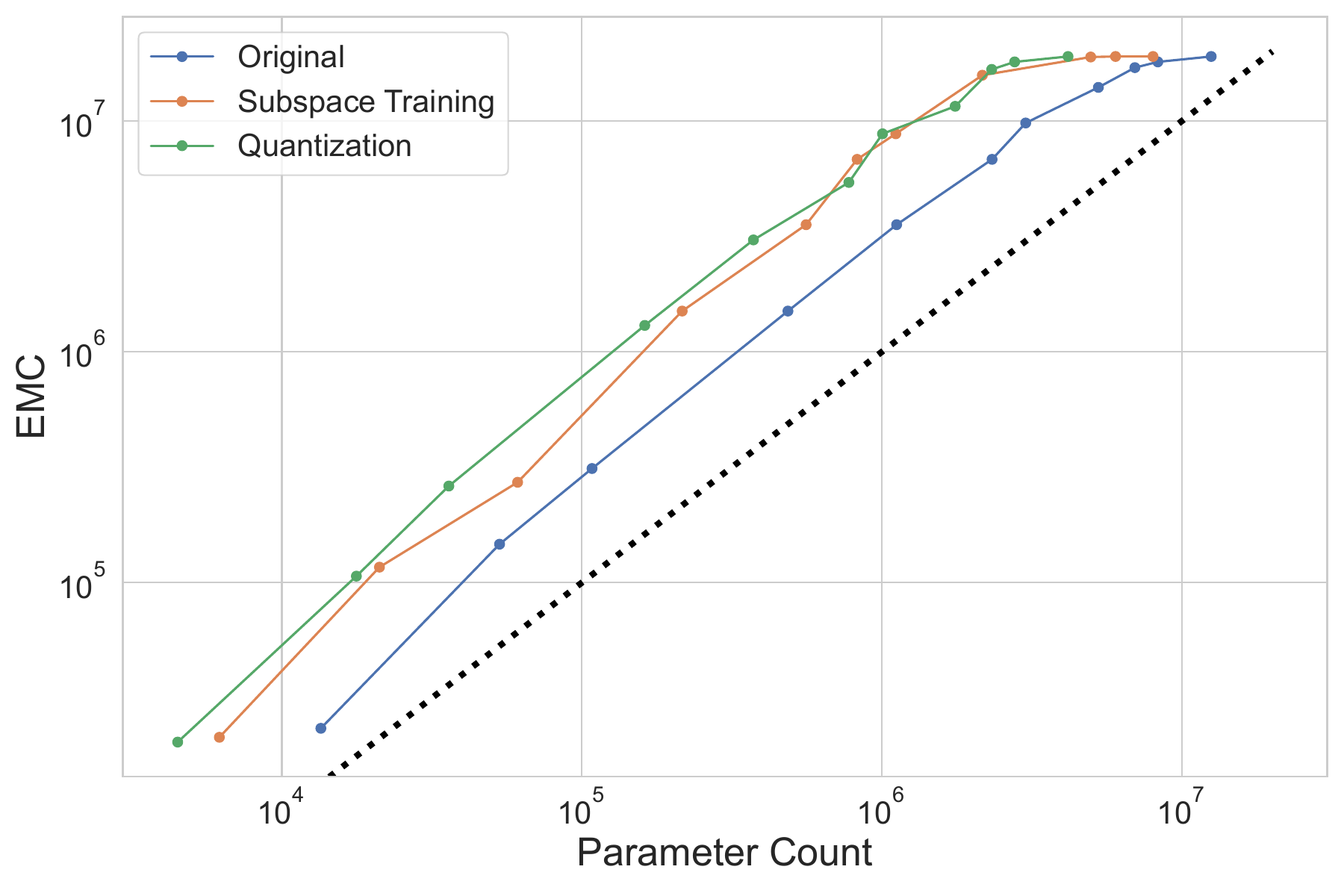}
        \caption{Original Labels}
        \label{fig:subb2}
    \end{subfigure}
    \hfill
     \begin{subfigure}{0.48\textwidth}
        \centering
        \includegraphics[width=1\linewidth]{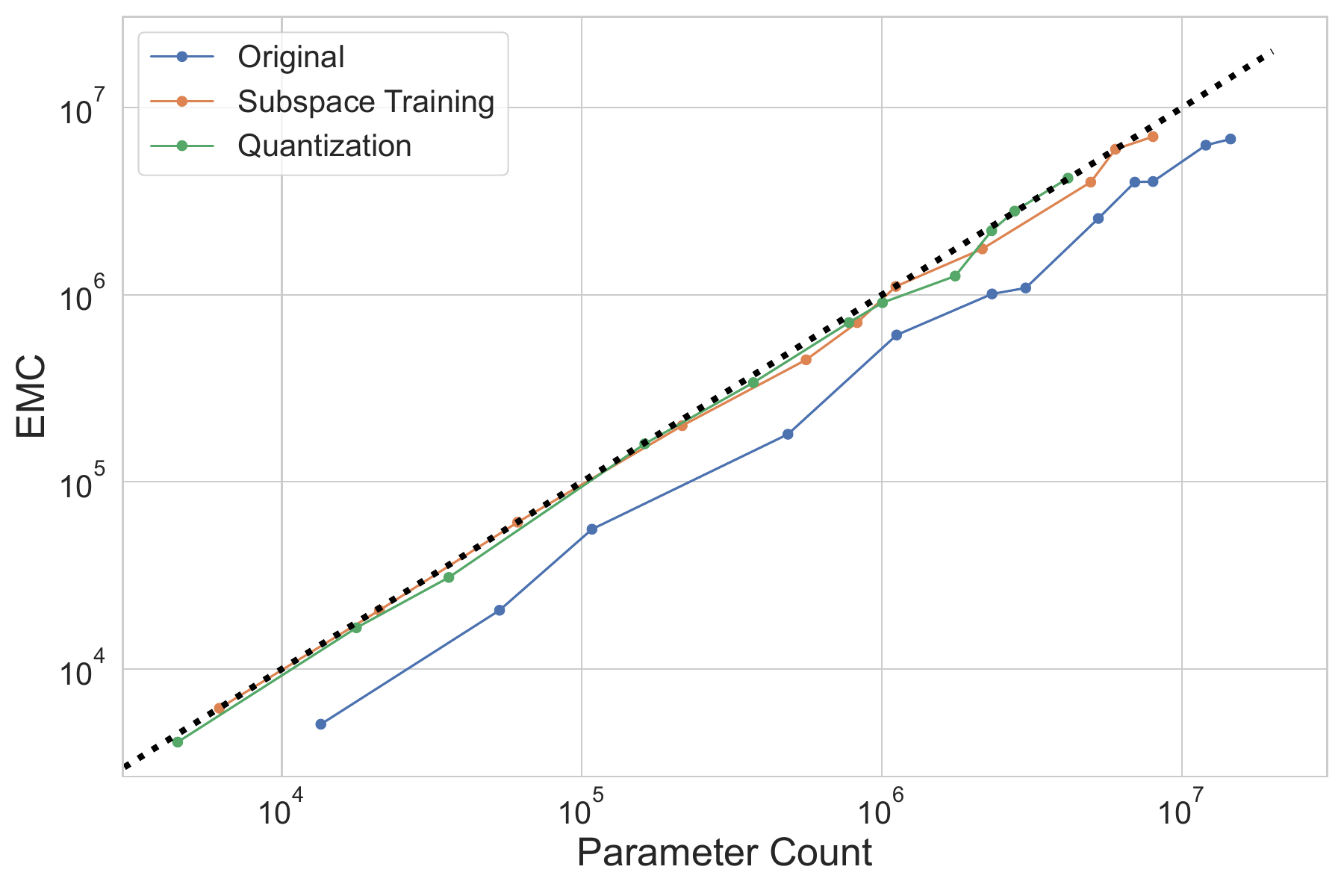}
        \caption{Random Labels}
        \label{fig:suba3}
    \end{subfigure}
    \caption{\textbf{Compression improves Network efficiency.} The EMC across different model sizes for original and random labels.  CNN architectures on ImageNet-20MS.}
    \label{fig:apend_compression}
\end{figure}

\begin{figure}[H]
\centering
\includegraphics[width=1\linewidth]{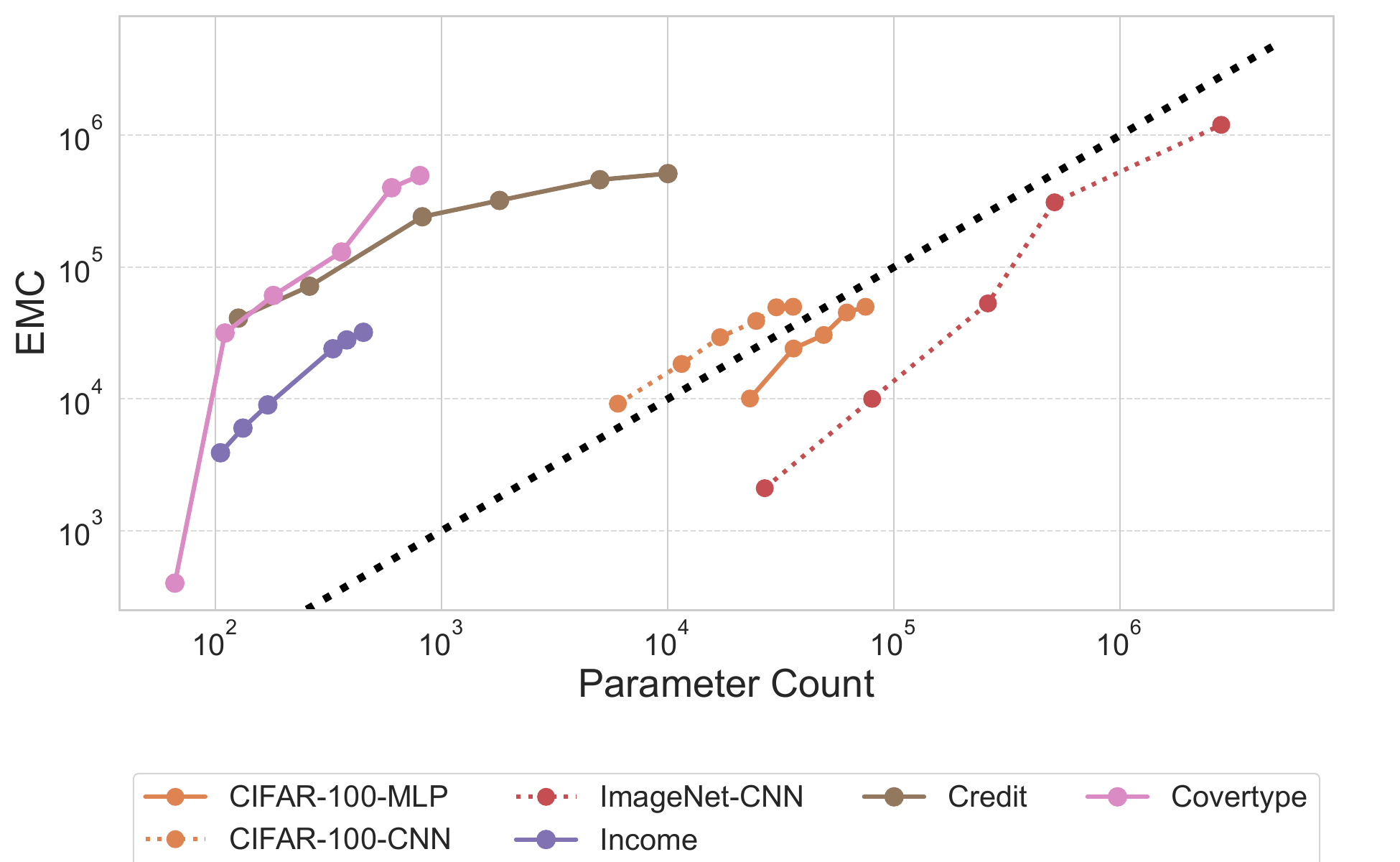}
\caption{\textbf{EMC as function of the number of parameters for datasets that converted to binary classification.}}
\label{fig:binary_datasets}
\end{figure}

\subsection{Implementation Details}
\label{app:impl_details}
Unless otherwise mentioned, our hyperparameter tuning was conducted over the following hyperparameters: batch size - with the values \([32, 64, 128, 256]\). For the Stochastic Gradient Descent (SGD) optimizer, we used an initial learning rate selected by grid search between \(0.001\) and \(0.01\) with Cosine annealing. For Adam and AdamW optimizers, the learning rate was chosen by grid search between \(1e-5\) and \(1e-2\).

For other hyperparameters, we adhere to the standard PyTorch recipes.

\subsection{Empirical Model Complexity }
\label{ref:edc}

To compute the Empirical Model Complexity (EMC), we adopt an iterative approach for each network size. Initially, we start with a small number of samples and train the model. Post-training, we verify if the model has perfectly fit all the samples by achieving $100\%$ training accuracy. If this criterion is met, we re-initialize the model with a random initialization and train it again on a larger number of samples, randomly drawn from the full dataset. This process is iteratively performed, increasing the number of samples in each iteration, until the model fails to perfectly fit all the training samples. The largest sample size where the model achieves a perfect fit is taken as the Empirical Model Complexity for that particular network size. It is important to note that data is sampled independently on each iteration.

While it is possible to artificially prevent models from fitting their training set by under-training, thus confounding any study of capacity to fit data, we ensure that all training runs reach a minimum of the loss function by imposing three conditions: 

First, the norm of the gradients across all samples must fall below a pre-defined threshold. We observed that there is a high variance in the norms of the gradients between different networks; therefore, we set this threshold manually after checking the norms for each network type when training with a small number of samples, where it's clear that the networks fit perfectly and converge to a minimum.

Second, the training loss should stabilize. To ensure this, we stipulate that the average loss should not decrease for 10 consecutive epochs.

Third, we check for the absence of negative eigenvalues in the loss Hessian to confirm that the model has indeed reached a minimum rather than a saddle point. To do this, we calculate the eigenvalues using the PyHessian Python package \citep{yao2020pyhessian} and validate that after training converges, there are no eigenvalues smaller than $-1e-2$. This threshold was chosen after examining the eigenvalue distributions of different networks that fit perfectly.

\subsection{Compute Resources}
\label{ref:compute}

Our experiments were conducted using NVIDIA Tesla V100 GPUs with 32GB memory each for model training and evaluation. The total compute time for the entire set of experiments was approximately 3000 GPU hours. All experiments were run on NUY's cluster managed with SLURM, ensuring efficient resource allocation and job scheduling. This setup allowed us to handle the extensive computational demands of training large neural network models and conducting comprehensive evaluations.

\section{Broader Impacts}
\label{ref:baroder}

Our research on the capacity of neural networks to fit data more efficiently has several important implications. Positively, our findings could lead to more efficient AI models, which would benefit various applications by making these technologies more accessible and effective. By understanding how neural networks can be more efficient, we can also reduce the environmental impact associated with training large models.

However, there are potential negative impacts as well. Improved neural network capabilities might be used in ways that invade privacy, such as through enhanced surveillance or unauthorized data analysis. Additionally, as AI technologies become more powerful, it is essential to consider ethical implications, fairness, and potential biases in their development and use.

To address these concerns, our paper emphasizes the importance of responsible AI practices. We encourage transparency, ethical considerations, and ongoing research into the societal impacts of advanced machine learning technologies to ensure they are used for the greater good.

\section{Limitations}
\label{appendix:limitations}

Our study has several limitations that should be considered when interpreting the results. First, the datasets used in our experiments, while diverse, may not fully represent the wide variety of data encountered in practical applications. This could introduce biases and limit the generalizability of our findings. Second, our experiments are constrained by the available computational resources. While we used NVIDIA Tesla V100 GPUs with 32GB memory, the total compute time was approximately 3000 GPU hours. This limitation restricted the scale and number of experiments we could perform, potentially affecting the robustness of our conclusions.

Furthermore, our analysis primarily focuses on certain types of neural network architectures, such as CNNs, MLPs, and ViTs. While these are common and widely used, there are many other architectures that we did not explore. The impact of different training procedures, regularization techniques, and hyperparameter choices on the EMC might vary with other architectures.

Additionally, we decided to test a wide range of factors affecting neural network flexibility but only explored a limited number of settings for each factor, rather than delving deeply into any single factor. This broad but shallow exploration might miss deeper insights that a more focused study could reveal.

Lastly, our method of measuring EMC, while rigorous, relies on specific criteria for determining when a model has perfectly fit its training data. These criteria include achieving 100\% training accuracy and the absence of negative eigenvalues in the loss Hessian. Different criteria might yield slightly different EMC values, and this should be taken into account when applying our findings to other contexts.

Despite these limitations, we believe our study provides valuable insights into the factors influencing neural network flexibility and highlights areas for further research.

\end{document}

%% file: math_commands.tex
\usepackage{amsmath,amsfonts,bm}

\def\eqref#1{equation~\ref{#1}}
\def\1{\bm{1}}

\DeclareMathAlphabet{\mathsfit}{\encodingdefault}{\sfdefault}{m}{sl}
\SetMathAlphabet{\mathsfit}{bold}{\encodingdefault}{\sfdefault}{bx}{n}